\documentclass[10pt,journal,compsoc]{IEEEtran}

\ifCLASSOPTIONcompsoc
  \usepackage[nocompress]{cite}
\else
  \usepackage{cite}
\fi

\ifCLASSINFOpdf
\else
\fi

\usepackage[utf8]{inputenc}
\usepackage{amsmath}
\usepackage{graphicx}
\usepackage{xspace}
\usepackage{multirow}
\usepackage{multicol}
\usepackage{booktabs}
\usepackage{color}
\usepackage{xcolor}
\usepackage{colortbl}
\usepackage{amssymb}
\usepackage{subfigure}
\usepackage{algorithm}
\usepackage{algorithmic}

\title{Out-of-distribution Evidence-aware Fake News Detection via Dual Adversarial Debiasing}

\author{Qiang Liu, \IEEEmembership{Member, IEEE},
        Junfei Wu,
        Shu Wu, \IEEEmembership{Senior Member, IEEE},
        and Liang Wang, \IEEEmembership{Fellow, IEEE}
\IEEEcompsocitemizethanks{
\IEEEcompsocthanksitem All authors are with the Center for Research on Intelligent Perception and Computing (CRIPAC), State Key Laboratory of Multimodal Artificial Intelligence Systems (MAIS), Institute of Automation, Chinese Academy of Sciences (CASIA), and are also with the School of Artificial Intelligence, University of Chinese Academy of Sciences, Beijing, China.\protect\\
E-mail: qiang.liu@nlpr.ia.ac.cn, junfei.wu@cripac.ia.ac.cn, \{shu.wu, wangliang\}@nlpr.ia.ac.cn.
}
\thanks{
(Corresponding author: Shu Wu)}}
\newcommand{\themodel}{DAL\xspace}

\begin{document}

\IEEEtitleabstractindextext{%
\begin{abstract}
Evidence-aware fake news detection aims to conduct reasoning between news and evidences, which are retrieved based on news content, to find uniformity or inconsistency.
However, we find evidence-aware detection models suffer from biases, i.e., spurious correlations between news/evidence contents and true/fake news labels, and are hard to be generalized to Out-Of-Distribution (OOD) situations.
To deal with this, we propose a novel Dual Adversarial Learning (DAL) approach.
We incorporate news-aspect and evidence-aspect debiasing discriminators, whose targets are both true/fake news labels, in DAL.
Then, DAL reversely optimizes news-aspect and evidence-aspect debiasing discriminators to mitigate the impact of news and evidence content biases.
At the same time, DAL also optimizes the main fake news predictor, so that the news-evidence interaction module can be learned.
This process allows us to teach evidence-aware fake news detection models to better conduct news-evidence reasoning, and minimize the impact of content biases.
To be noted, our proposed DAL approach is a plug-and-play module that works well with existing backbones.
We conduct comprehensive experiments under two OOD settings, and plug DAL in four evidence-aware fake news detection backbones.
Results demonstrate that, DAL significantly and stably outperforms the original backbones and some competitive debiasing methods.
\end{abstract}

}

\maketitle


\section{Introduction}  \label{sec:intro}

\IEEEPARstart{W}{ith} the development of online media, users can access to information more easily, and messages can be spread more rapidly.
However, at the same time, fake statement about news can also be spread to the public more easily and widely.
This leads to potential harm to the society, and may cause severe consequences.
For example, rumors about Covid-19 have seriously affected the public health \cite{Naeem2020TheC}.
Thus, it is necessary to conduct research on automatic fake news detection \cite{castillo2011information}, which remains a challenging task.

In the past decade, the task of fake news detection has been widely studied from different perspectives.
Most research works focus on extracting and modeling patterns between true news and fake news based on different types of related features, including textual contents \cite{volkova2017separating,giachanou2019leveraging,przybyla2020capturing}, multimodal contents \cite{jin2017multimodal,khattar2019mvae,qian2021hierarchical} and propagation structures \cite{kwon2013prominent,ma2015detect,ma2016detecting,bian2020rumor}, to detect fake news on online media.
On the other hand, evidence-aware fake news detection \cite{popat2017truth,popat2018declare,vo2021hierarchical,xu2022evidence} aims to conduct textual reasoning between news, which is the claim of a fact, and evidences, which are retrieved from news platforms according to the content of the news.
With the help of evidences, compared with other types of models, evidence-aware fake news detection models are more reliable and interpretable.
In this work, we focus on the evidence-aware fake news detection task.

In real-world fake news detection systems, we usually face the Out-Of-Distribution (OOD) problem \cite{hansen2021automatic,lin2022detect,wang2018eann,choi2021using}, in which training phase and testing phase share different data distributions.
We usually train fake news detection models on data from limited platforms, and need to apply them to generalized platforms.
Meanwhile, during training of fake news detection models, we usually only have data from limited topics or events in constrained time periods, and require the models to generalize to more environments.

As datasets for training are usually with biased data distribution, evidence-aware fake news detection models may mistakenly learn spurious correlations between news contents and true/fake news labels.
Considering the content relevance between news and evidences, spurious correlations between evidence contents and true/fake news labels \cite{hansen2021automatic} may also be captured.
These spurious correlations bring evidence-aware fake news detection models severe OOD problems.
We denote above biases as \textbf{news content bias} and \textbf{evidence content bias} respectively.
Both biases limit evidence-aware fake news detection models to well reason between news and evidences.
For example, if training data contains news about an accident event which is mostly fake news, the models have chance to learn that the topics and key-words about the accident refer to fake news, instead of conducting news-evidence reasoning. 
Obviously, above learned correlations do not hold when data distribution changes.
Meanwhile, when training data and testing data share different topic distributions, e.g., entertainment and politics, correlations between some key-words in the training data and true/fake news labels may be captured, which may not exist during the testing phase.
Moreover, in Sec. \ref{sec:empirical}, with empirical experiments, we prove that existing evidence-aware models have trouble in OOD environments.
Accordingly, it is necessary to mitigate the impact of news content bias and evidence content bias, and better teach detection models to conduct news-evidence reasoning.

However, the OOD problem has not been sufficiently studied in fake news detection, especially for the evidence-aware task \cite{hansen2021automatic}.
Some approaches investigate multi-domain detection \cite{nan2021mdfend,zhu2022memory} or cross-domain detection \cite{silva2021embracing,nan2022improving,lin2022detect}, but still require observations in target domains.
Some multimodal fake news detection models \cite{wang2018eann,choi2021using} focus on the situation that training samples and testing samples come from different events.
However, they only focus on common features among different events or topics, but neglect the specific reasoning path in evidence-aware fake news detection.
Recently, counterfactual inference \cite{niu2021counterfactual} has been applied to debiasing evidence-aware fake news detection models \cite{wu2022bias}.
However, it is performed during the testing phase, and hard to be adaptively optimized.
Meanwhile, there is a similar task of evidence-aware fake news detection called fact verification \cite{thorne2018fever,zhou2019gear,liu2020fine}.
Some efforts have made for debiasing fact verification models \cite{lee2021crossaug,schuster2019towards}.
However, biases in fact verification are different from those in evidence-aware fake news detection.
The fact verification datasets are usually human-annotated, and negative samples are generated via data augmentation in claims by adding words such as ''not."
Thus, the biases in fact verification are the spurious correlations between some negative words in claims and labels.
On the other hand, biases in evidence-aware fake news detection come from some topics and events in both news and evidence contents.

In this paper, we aim to mitigate both news and evidence content biases, and obtain detection models with great OOD generalization ability.
Inspired by domain-adversarial training \cite{ganin2015unsupervised,ganin2016domain}, we propose a novel \textbf{Dual Adversarial Learning (DAL)} approach, for debiasing evidence-aware fake news detection models.
The proposed DAL approach is a plug-and-play module, which can be applied in various evidence-aware fake news detection models.
(1) For content of a piece of news, we conduct mean pooling on word embeddings of the news extracted in a detection model with no evidence content information, and obtain news representation.
For content of each retrieved evidence, similarly, we conduct mean pooling on word embeddings of the evidence extracted in a detection model with no news content information, and obtain evidence representation.
(2) In dual aspects, we remove the spurious correlations between news/evidence contents and true/fake news labels.
Specifically, we use the news representation and evidence representations to construct news-aspect and evidence-aspect debiasing discriminators respectively, via simple multi-layer perception networks.
(3) We conduct word-level and sentence-level interaction between news representation and evidence representations, as existing evidence-aware models do, to construct a main fake news predictor.
(4) Simultaneously, we positively optimize the main fake news predictor, while reversely optimize the news-aspect and evidence-aspect debiasing discriminators.
In this way, during the learning of the evidence-aware fake news detector, we can mitigate the spurious correlations between news/evidence contents and true/fake news labels as much as possible.

Furthermore, we conduct experiments under two OOD settings, i.e., cross-platform and cross-topic, and plug the proposed DAL approach in four evidence-aware fake news detection backbones.
Our approach can significantly and stably outperform the original detection backbones and several state-of-the-art debiasing baselines, which shows the effectiveness of DAL for debiasing evidence-aware fake news detection models in OOD environments.

Our main contributions can be summarized as follows:
\begin{itemize}
    \item We introduce news content bias and evidence content bias in evidence-aware fake news detection, and propose to mitigate them for training detection models with better OOD generalizing ability.
    \item We propose a plug-and-play dual adversarial learning approach, which incorporates both news-aspect debiasing and evidence-aspect debiasing modules.
    \item Comprehensive experiments are conducted to demonstrate the superiority of DAL in different OOD environments, and promote existing evidence-aware fake news detection backbones.
\end{itemize}

The rest of the paper is organized as follows.
In Sec. 2, we review some related work on fake news detection and debiasing methods.
Then, in Sec. 3, we present causal view analysis, and then conduct some empirical experiments to show the performances of existing evidence-aware models in OOD settings.
Sec. 4 details our proposed DAL approach for debiasing evidence-aware fake news detection models.
In Sec. 5, we conduct empirical experiments to verify the effectiveness of DAL.
Finally, Sec. 6 concludes our work.

\section{related work} \label{sec:related_work}

In this section, we first briefly review some related works on different types of fake news detection tasks, i.e., content-based, pattern-based and evidence-aware.
Then, we introduce some debiasing methods which have been used in fake news detection tasks.

\subsection{Content-based Fake News Detection}

Content-based fake news detection solely relies on content features for identifying misinformation.
The considered content features can be grouped into two categories: textual features and multimodal features.

Fake news detection models based on textual features aim to find linguistic patterns of true/fake news \cite{volkova2017separating,rashkin2017truth}.
Existing research works on text-based fake news detection attempt to analyze style \cite{przybyla2020capturing} or emotion \cite{giachanou2019leveraging,zhang2021mining} of news.
Meanwhile, BERT \cite{kenton2019bert} has been widely used for detecting fake news recently \cite{kaliyar2021fakebert}.
And the NEP model \cite{sheng2022zoom} investigates popularity and novelty of news to detect misinformation from macro and micro environments respectively.

On the other hand, multimodal fake news detection considers both texts and images of news \cite{jin2016novel,jin2017multimodal} for identifying misinformation.
The major problem to be investigated and solved is the interaction and fusion among multimodal features, and methods such as recurrent neural networks \cite{jin2017multimodal}, multimodal variational autoencoder \cite{khattar2019mvae} and multimodal attention \cite{qian2021hierarchical} have been applied.
Moreover, some works focus discovering the inconsistency among different modalities for multimodal fake news detection \cite{tan2020detecting,qi2021improving,chen2022cross}.

\subsection{Propagation-based Fake News Detection}

Propagation-based fake news detection aims to identify misinformation with feedback in online social media, such as reposts, likes, and comments \cite{kwon2013prominent,ma2015detect}.
The core problem of propagation-based fake news detection is to aggregate the propagation history of a piece of news, and plenty of works have been published.
Among them, models based on recurrent neural networks \cite{ma2016detecting,ma2018rumor}, convolutional neural networks \cite{yu2017convolutional,liu2018early}, attentive networks \cite{liu2018mining}, and generative adversarial networks \cite{ma2019detect} have been proposed.

With the development of Graph Neural Network (GNN) \cite{kipf2017semi}, recent works on propagation-based fake news detection mainly model propagation structures as graphs and applies GNN for misinformation identification \cite{bian2020rumor,lu2020gcan,nguyen2020fang}.
Furthermore, some works based on contrastive learning \cite{sun2022rumor} or hypergraph \cite{sun2022structure} have been further proposed.
For better interpretability, reasoning over subgraph has been investigated \cite{jin2022towards,yang2022reinforcement}.
Meanwhile, for capturing temporal characteristics, dynamic graph has been constructed and applied for rumor detection \cite{sun2022ddgcn}.

\subsection{Evidence-aware Fake News Detection}

Evidence-aware fake news detection seeks to explore the semantic similarity or conflict between news and related evidences to identify misinformation \cite{popat2017truth}.
DeClare \cite{popat2018declare} employs BiLSTMs to embed the news and evidences.
Then, it computes news representation via mean pooling, and computes news-aware attention for each word in evidences.
HAN \cite{ma2019sentence} computes the sentence-level coherence and entailment scores between news and evidences.
EHIAN \cite{wu2021evidenceIJCAI} incorporates self-attention for word-level interaction.
MAC \cite{vo2021hierarchical} proposes hierarchical mutli-head attention networks to model interactions between news and evidences.
Then, some works propose to hierarchically conduct word-level and sentence-level interactions \cite{wu2021evidenceAAAI,wu2021unified}.
Recently, GET \cite{xu2022evidence} exploits graph-structure reasoning between news content and evidence contents, and GETRAL \cite{wu2022adversarial} further incorporates contrastive learning for representation leaning of news and evidences.
Moreover, the way integrating pattern-based and evidence-aware fake news detection has also been investigated \cite{sheng2021integrating}.
Meanwhile, there is research showing that, evidence-aware fake news detection models can hardly learn about news-evidence reasoning, but capture spurious correlations between news/evidence contents and true/fake news labels \cite{hansen2021automatic}.

\subsection{Debiasing Methods for Fake News Detection}

Some research works have tried to debias different types of fake news detection models or models in some related tasks.

The causal theory has been applied for debiasing fake news detection models or fact checking models.
Under the causal intervention framework \cite{schnabel2016recommendations,zhang2021causal,hu2022causal}, the sample weighting strategy \cite{clark2019don}, which downsamples the contribution of biased samples during loss computation, has been adopted for debiasing fake news detection models.
For example, PeE \cite{mahabadi2020end} constructs a bias-only model, and downsamples the samples' spurious class distribution.
For debiasing fact verification models, ReW \cite{schuster2019towards} downsamples the samples containing n-grams highly correlated with labels.
Meanwhile, counterfactual inference \cite{niu2021counterfactual}, which subtracts the effects of spurious correlations estimated by a bias-only model, is applied in eliminating entity bias \cite{zhu2022generalizing}, or debiasing evidence-aware fake news detection models \cite{wu2022bias} and fact checking models \cite{xu2023counterfactual}.

Besides, data augmentation strategies aim to generate some unbiased samples and incorporate them for training \cite{wei2019eda}.
For the fact verification task, CrossAug \cite{lee2021crossaug} proposes a cross contrastive augmentation strategy, in which original claim text is modified to be negative, and evidences are changed to support the modified claim, to deal with the spurious correlations between some negative words in claim and labels.
Recently, prompt learning has also been used for dealing with cross-language and cross-domain problems in propagation-based fake news detection \cite{lin2023zero}.
Moreover, inspired by domain-adversarial training  strategy \cite{ganin2015unsupervised,ganin2016domain}, EANN \cite{wang2018eann} incorporates an event discriminator to mitigate the correlations between input features and domain information.
And similar strategy is used for fake news video detection \cite{choi2021using}.
Compared with our proposed DAL approach, EANN directly applies the domain-adversarial training strategy for dealing with the general domain shift problem.
In contrast, DAL deeply investigates the specific biases in evidence-aware fake news detection models and designs suitable adversarial losses for mitigating them.

\begin{figure}
	\centering
	\subfigure[Causal diagram of existing detection models.]{
		\begin{minipage}[b]{0.23\textwidth}
			\includegraphics[width=1\textwidth]{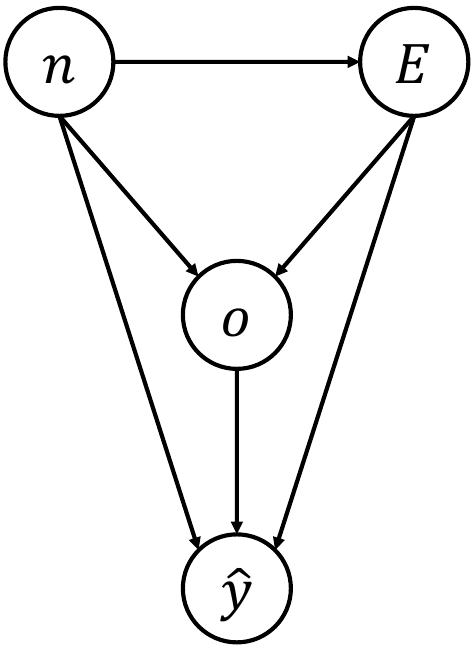}
		\end{minipage}
	\label{fig:causal2}
	}
	\subfigure[Removing spurious correlations $n \to \hat{y}$ and $E \to \hat{y}$.]{
		\begin{minipage}[b]{0.23\textwidth}
			\includegraphics[width=1\textwidth]{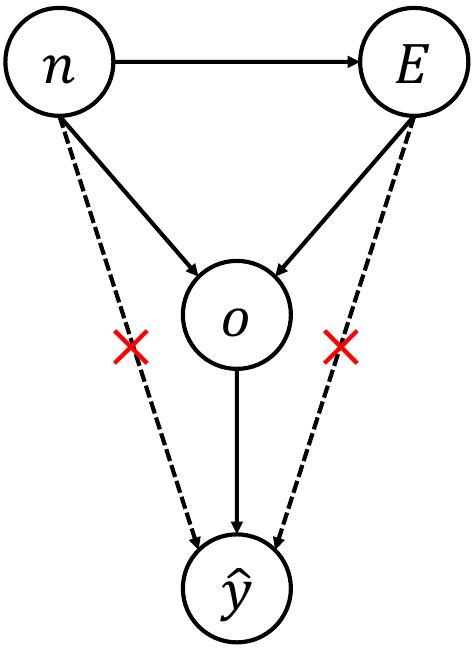}
		\end{minipage}
	\label{fig:causal3}
	}
	\caption{Causal diagrams of evidence-aware fake news detection.}
	\label{fig:causal}
\end{figure}

\begin{table*}[htbp]
  \centering
  \caption{Empirical experimental results, in which training and testing are conducted on the same dataset with random splitting.}
    \begin{tabular}{cccccc}
    \toprule
    \multirow{2}[2]{*}{Backbone} & \multirow{2}[2]{*}{Input} & \multicolumn{2}{c}{Training: PolitiFact, Testing: PolitiFact} & \multicolumn{2}{c}{Training: Snopes, Testing: Snopes} \\
          &       & F1-Macro & F1-Micro & F1-Macro & F1-Micro \\
    \midrule
    \multirow{3}[2]{*}{BERT} & News  & 0.6258  & 0.6367  & 0.6232  & 0.6867  \\
          & Envidences & 0.6413  & 0.6433  & 0.6598  & 0.7235  \\
          & News+Envidences & 0.6529  & 0.6624  & 0.6709  & 0.7481  \\
    \midrule
    \multirow{3}[2]{*}{DeClare} & News  & 0.6251  & 0.6303  & 0.6159  & 0.6634  \\
          & Envidences & 0.6392  & 0.6440  & 0.6522  & 0.7305  \\
          & News+Envidences & 0.6508  & 0.6590  & 0.6642  & 0.7572  \\
    \midrule
    \multirow{3}[2]{*}{MAC} & News  & 0.6244  & 0.6344  & 0.6090  & 0.6703  \\
          & Envidences & 0.6455  & 0.6465  & 0.6637  & 0.7167  \\
          & News+Envidences & 0.6609  & 0.6642  & 0.6725  & 0.7552  \\
    \midrule
    \multirow{3}[2]{*}{GET} & News  & 0.6299  & 0.6329  & 0.6421  & 0.6999  \\
          & Envidences & 0.6298  & 0.6344  & 0.6610  & 0.7325  \\
          & News+Envidences & 0.6567  & 0.6702  & 0.6741  & 0.7545  \\
    \bottomrule
    \end{tabular}%
  \label{tab:pilot1}%
\end{table*}%

\begin{table*}[htbp]
  \centering
  \caption{Empirical experimental results, in which training and testing are conducted on different datasets.}
    \begin{tabular}{cccccc}
    \toprule
    \multirow{2}[2]{*}{Backbone} & \multirow{2}[2]{*}{Input} & \multicolumn{2}{c}{Training: Snopes, Testing: PolitiFact} & \multicolumn{2}{c}{Training: PolitiFact, Testing: Snopes} \\
          &       & F1-Macro & F1-Micro & F1-Macro & F1-Micro \\
    \midrule
    \multirow{3}[2]{*}{BERT} & News  & 0.5112  & 0.5050  & 0.5353  & 0.6074  \\
          & Envidences & 0.5003  & 0.5099  & 0.5437  & 0.6094  \\
          & News+Envidences & 0.4932  & 0.5147  & 0.5569  & 0.6204  \\
    \midrule
    \multirow{3}[2]{*}{DeClare} & News  & 0.5372  & 0.5372  & 0.5313  & 0.6250  \\
          & Envidences & 0.5265  & 0.5265  & 0.5267  & 0.6214  \\
          & News+Envidences & 0.5328  & 0.5350  & 0.5488  & 0.6442  \\
    \midrule
    \multirow{3}[2]{*}{MAC} & News  & 0.5256  & 0.5287  & 0.5277  & 0.6115  \\
          & Envidences & 0.5414  & 0.5427  & 0.5361  & 0.6036  \\
          & News+Envidences & 0.5484  & 0.5497  & 0.5443  & 0.6320  \\
    \midrule
    \multirow{3}[2]{*}{GET} & News  & 0.5359  & 0.5420  & 0.5307  & 0.6486  \\
          & Envidences & 0.5265  & 0.5265  & 0.5321  & 0.6384  \\
          & News+Envidences & 0.5473  & 0.5593  & 0.5420  & 0.6581  \\
    \bottomrule
    \end{tabular}%
  \label{tab:pilot2}%
\end{table*}%

\section{Analysis} \label{sec:pre}

In this section, we present a causal view analysis of evidence-aware fake news detection models, and then conduct some empirical experiments to show the performances of existing detection models in OOD settings.

\subsection{Causal Diagrams}  \label{sec:analysis}

We have news $n$, and the corresponding evidences $E = \{ {e_0 , e_1 , e_2 ,...} \}$.
Via news-evidence interaction, we obtain interaction feature $o$, and the final prediction $\hat y$ on true/fake news label.
As in some causal inference-based debiasing approaches, we rely on causal diagrams for investigating the specific biases in evidence-aware fake news detection models and correspondingly designing debiasing strategies.
In Fig. \ref{fig:causal}, we illustrate the causal diagrams of evidence-aware fake news detection.

Evidence-aware fake news detection models usually conduct interaction between news and evidences and obtain interaction feature, i.e., $n \to o$ and $E \to o$.
We usually wish the models can well perform news-evidence reasoning and give predictions, i.e., $o \to \hat y$.
However, due to biases in data, the models may learn the spurious correlation between news content and true/fake news labels, i.e., $n \to \hat{y}$, without conducting new-evidence reasoning.
Meanwhile, considering evidences are retrieved according to news contents, contents of news and evidences are highly correlated, and they may share similar topics or key-words, i.e., $n \to E$.
Due to biased data distribution, these topics and key-words may be mistakenly associated with true/fake labels, which results in the spurious correlation between evidence content and true/fake news labels, i.e., $E \to \hat{y}$.
These spurious correlations may change cross different OOD environments, in which we usually have $\mathop p\nolimits_{{\rm{train}}} \left( {\hat y|n} \right) \ne \mathop p\nolimits_{{\rm{test}}} \left( {\hat y|n} \right)$ and $\mathop p\nolimits_{{\rm{train}}} \left( {\hat y|E} \right) \ne \mathop p\nolimits_{{\rm{test}}} \left( {\hat y|E} \right)$.
Examples have been already discussed in Sec. \ref{sec:intro}.
Thus, it is necessary to mitigate the impact of spurious correlations $n \to \hat{y}$ and $E \to \hat{y}$, so that evidence-aware fake news detection models are able to better reason between news and evidences.

\subsection{Empirical Experiments} \label{sec:empirical}

In \cite{hansen2021automatic}, the authors claim that evidence-aware fake detection models can hardly conduct news-evidence reasoning, but merely capture biases in contents.
To clarify our motivation, we conduct further empirical experiments.
We consider two datasets, i.e., PoliticFact and Snopes \cite{hansen2021automatic}, and four detection models, i.e., BERT \cite{kenton2019bert}, DeClare \cite{popat2018declare}, MAC \cite{vo2021hierarchical} and GET \cite{xu2022evidence}.
In Tab. \ref{tab:pilot1}, training and testing are conducted on the same platform, i.e. dataset.
Meanwhile, in Tab. \ref{tab:pilot2}, we conduct cross-platform training and testing.
We have three types of input features to the models: only news, only evidences, and both news and evidences.
Obviously, performances with both features are very close to those with only news features or evidence features.
Moreover, comparing results in Tab. \ref{tab:pilot1} and results in Tab. \ref{tab:pilot2}, we can conclude that, the out-of-distribution problem severely affects the performances of evidence-aware fake news detection models.
Accordingly, we need to teach models to better conduct news-evidence reasoning, and mitigate news and evidence content biases.

\section{Methodology}
\label{sec:method}

\begin{figure*}
    \centering
    \includegraphics[width=0.95\linewidth]{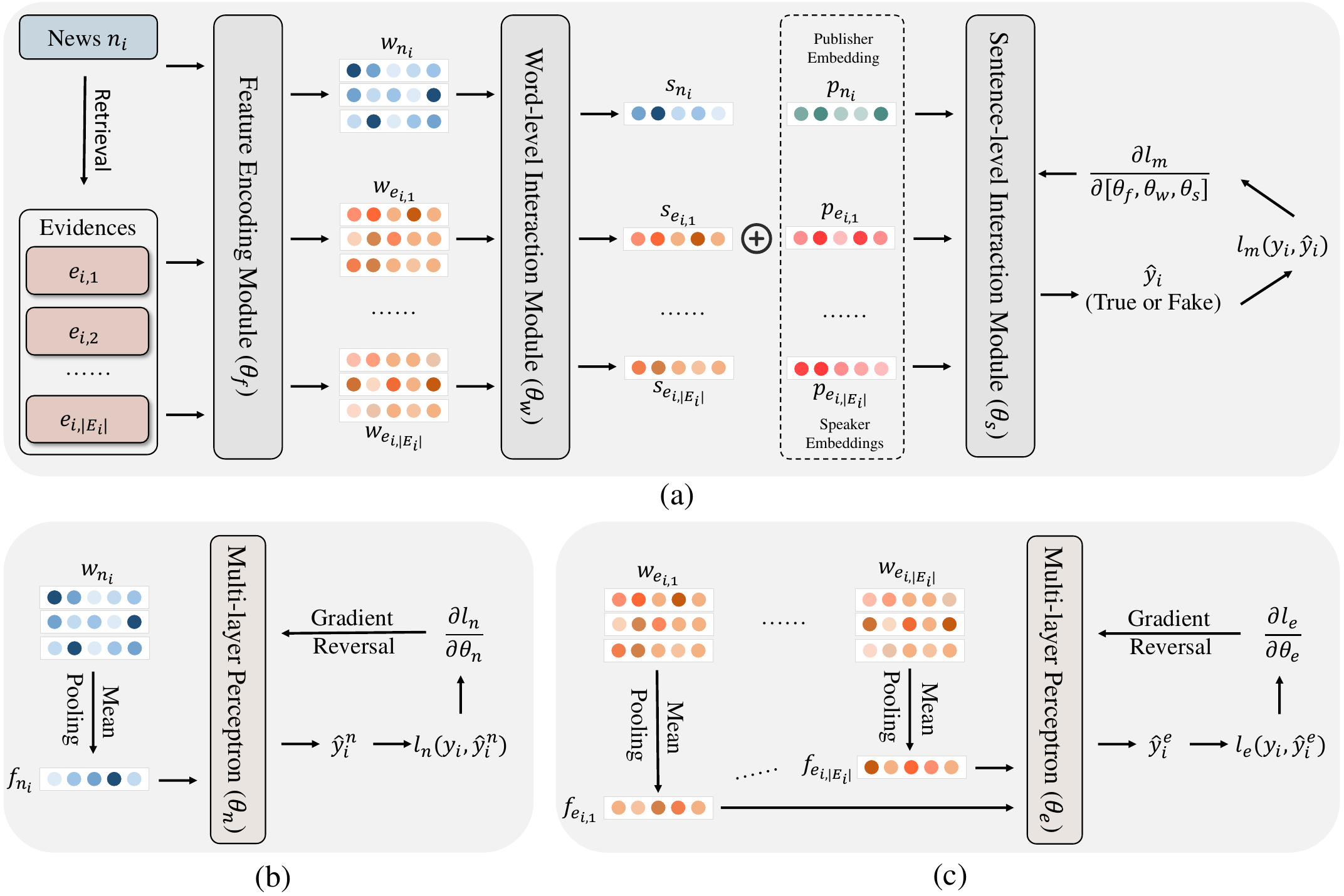}
    \caption{The overview of DAL with a piece of news $n_i$ and its corresponding evidences $E_i  = \left\{ {e_{i,1} , e_{i,2} ,..., e_{i,\left| {E_i } \right|} } \right\}$: (a) the structure of evidence-aware fake news detection models; (b) the structure of the news-aspect debiasing; (c) the structure of the evidence-aspect debiasing.}
    \label{fig:model}
\end{figure*}

In this section, we detail our proposed DAL approach.
We start with problem formulation of evidence-aware fake news detection.
Then, we introduce the common structure of existing evidence-aware fake news detection models.
Finally, we introduce our dual adversarial debiasing strategy.

\subsection{Problem Formulation}
\label{task}

Evidence-aware fake news detection is a binary classification task, in which the detection model aims to predict the probability of true/fake news.
We have a set of news to be verified denoted as $\mathcal{N} = \left\{ {n_1 , n_2 ,..., n_{\left| \mathcal{N} \right|} } \right\}$.
According to each news $n_i$, evidences are retrieved, and denoted as $E_i  = \left\{ {e_{i,1} , e_{i,2} ,..., e_{i,\left| {E_i } \right|} } \right\}$.
Respectively, $n_i$ and $e_{i,j}$ contain textual contents of the corresponding news and evidence.
The ground-truth true/fake news label of $n_i$ is denoted as $y_i \in \{ 0,1 \}$.
Based on $n_i$ and $E_i$, we need to give prediction $\hat y_i$ on veracity of the news.
To be noted, in this work, we focus on solving the OOD problem, in which training samples and testing samples are collected from different platforms, or share different topic distributions.

\subsection{Structure of Evidence-aware Fake News Detection}

In this subsection, we summarize the common structure of evidence-aware fake news detection models \cite{popat2017truth,popat2018declare,ma2019sentence,wu2021evidenceIJCAI,vo2021hierarchical,wu2021evidenceAAAI,wu2021unified,xu2022evidence}, which is shown in Fig. \ref{fig:model} (a).
With a piece of news $n_i$ and its corresponding retrieved evidences $E_i  = \left\{ {e_{i,1} , e_{i,2} ,..., e_{i,\left| {E_i } \right|} } \right\}$, a detection model firstly encodes input features into word-level embeddings as
\begin{equation}
    w_{n_i }  = Encoder\left( {n_i } \right),
\end{equation}
\begin{equation}
    w_{e_{i,j} }  = Encoder\left( {e_{i,j} } \right),
\end{equation}
where $w_{n_i }  \in \mathcal{R}^{\left| {n_i } \right| \times d_w }$ and $w_{e_{i,j} }  \in \mathcal{R}^{\left| {e_{i,j} } \right| \times d_w }$ are word-level embedding matrices for news and evidence respectively, and $d_w$ denotes the embedding dimensionality.
We denote the learnable parameters in the features encoder as $\theta_f$.

Then, the detection model conducts word-level news-evidence interaction, to generate sentence-level embeddings.
Sentence-level embeddings of news are usually obtained via variety of pooling operations as
\begin{equation}
    s_{n_i }  = Pooling\left( {w_{n_i } } \right),
\end{equation}
where $s_{n_i }  \in \mathcal{R}^{d_s }$, and $d_s$ denotes the embedding dimensionality.
And the interaction process for generating evidence embeddings can be formulated as
\begin{equation}
    s_{e_{i,j} }  = WorInter\left( {s_{n_i } , w_{ e_{i,j} } } \right),
\end{equation}
where $s_{e_{i,j} }  \in \mathcal{R}^{d_s }$.
This process conducts word-level interaction and reasoning between news and evidences.
We denote the learnable parameters in the word-level interaction module as $\theta_w$.

Finally, detection models further conduct sentence-level interaction between news and evidences for the final prediction on the true/fake news label.
This process can be formulated as
\begin{equation} \label{eq:sent_inter}
    \hat y_i  = SenInter\left( {s_{n_i } , s_{ e_{i,1} },...,s_{ e_{i,\left| {E_i } \right|} } } \right).
\end{equation}
We denote the learnable parameters in the sentence-level interaction module as $\theta_s$.
Moreover, in some detection models, publisher embedding and speaker embedding is involved to represent the credibility of authors of the news and corresponding evidences.
In this situation, Eq. (\ref{eq:sent_inter}) can be rewritten as
\begin{equation} \small
    \hat y_i  = SenInter\left( {s_{n_i }, p_{n_i }, s_{ e_{i,1} }, p_{ e_{i,1} },...,s_{ e_{i,\left| {E_i } \right|} } }, p_{ e_{i,\left| {E_i } \right|} }  \right),
\end{equation}
where $p_{n_i }$ and $p_{ e_{i,j} }$ denote publisher embedding and speaker embedding respectively.

We aim to propose a plug-and-play debiasing approach that can be applied for existing evidence-aware fake news detection models with feature encoding module, word-level interaction module and sentence-level interaction module as common components.
To make the structure of our approach clearer, we conclude several representative evidence-aware fake news detection models as follows.
HAN \cite{ma2019sentence} uses GRU \cite{dey2017gate} for feature encoding, adopts coherence and entailment attention for sentence-level interaction, and has no word-level interaction and publisher and speaker embeddings.
DeClare \cite{popat2018declare} uses BiLSTMs \cite{huang2015bidirectional} for feature encoding, conducts claim-aware attention for word-level interaction, involves publisher and speaker embeddings, and has no sentence-level interaction.
MAC \cite{vo2021hierarchical} uses BiLSTMs for feature encoding, multi-head attention for word-level interaction, multi-head attention for sentence-level interaction, and involves publisher and speaker embeddings.
GET \cite{xu2022evidence} incorporates graph neural networks \cite{kipf2016semi} for feature encoding, multi-head attention for word-level and sentence-level interaction, and involves publisher and speaker embeddings.

\subsection{News-aspect Debiasing Discriminator}

According to the analysis in Sec. \ref{sec:pre}, we need to remove the spurious correlations between news contents and labels.
To achieve this, we need firstly to estimate the dependency of the labels on news contents.
We obtain news representation via mean pooling as
\begin{equation}
    f_{n_i }  = MeanPooling\left( {w_{n_i } } \right),
\end{equation}
where $f_{n_i }  \in \mathcal{R}^{d_w }$.
Then, we incorporate a news-aspect debiasing predictor to estimate the dependency, with Multi-Layer Perceptron (MLP), as
\begin{equation}
    \hat y^n_i  = {MLP^n} \left( {f_{n_i } } \right).
\end{equation}
The learnable parameters in the discriminator are denoted as $\theta_n$.
With the discriminator, we can force the model not to accurately predict true/fake news labels based on only news contents, so that we can mitigate the news content bias.

\subsection{Evidence-aspect Debiasing Discriminator}

Considering the content similarity between news and evidences, as well as the analysis in Sec. \ref{sec:pre}, we also need to remove the spurious correlations between evidence contents and labels.
To ensure that evidence representations used for debiasing contain no information about the news, we take use of the word embeddings $w_{e_{i,j}}$ before news-evidence interactions.
Specifically, we also use mean pooling to generate the evidence representation as
\begin{equation}
    f_{e_{i,j} }  = MeanPooling\left( {w_{e_{i,j} } } \right),
\end{equation}
where ${f_{e_{i,j} }}  \in \mathcal{R}^{d_w }$.
Then, we incorporate another MLP as our evidence-aspect debiasing discriminator, to estimate the dependency of the labels on each evidence content.
The predictor can be formulated as the average of predictions by all the evidences as
\begin{equation}
    \hat y^e_i  = \frac{1}{\left| E_i \right|} \sum_{e_{i,j} \in E_i} {MLP^e} \left( {f_{e_{i,j} } } \right).
\end{equation}
The learnable parameters in the discriminator are denoted as $\theta_e$.
Similarly, with the discriminator, we are able to mitigate the evidence content bias.

\begin{algorithm}[t]
    \caption{Dual Adversarial Learning.}
    \label{alg:DAL}
    \begin{algorithmic}[1]
        \REQUIRE News set $\mathcal{N}$ and an evidence-aware model $M$.
        \ENSURE Model parameters $\theta_f$, $\theta_w$ and $\theta_s$.
        \STATE Initialize $k \gets 0$ and $k_{best} \gets 0$.
        \STATE Initialize $[\theta^{(0)}_f, \theta^{(0)}_w, \theta^{(0)}_s]$ in $M$, and $[\theta^{(0)}_n, \theta^{(0)}_e]$.
        \REPEAT
        \STATE Keep $[\theta^{(k)}_f, \theta^{(k)}_w, \theta^{(k)}_s]$ fixed, and update $[\theta^{(k+1)}_n, \theta^{(k+1)}_e]$ according to Eq. (\ref{eq:op_n}-\ref{eq:op_e}) on $\mathcal{N}$.
        \STATE Keep $[\theta^{(k+1)}_n, \theta^{(k+1)}_e]$ fixed, and update $[\theta^{(k+1)}_f, \theta^{(k+1)}_w,$ $\theta^{(k+1)}_s]$ according to Eq. (\ref{eq:op_m}) on $\mathcal{N}$.
        \STATE $k \gets k+1$.
        \STATE Update $k_{best} \gets k$, if better validation results reached.
        \UNTIL{Convergence.}
        \RETURN $\theta^{(k_{best})}_f$, $\theta^{(k_{best})}_w$ and $\theta^{(k_{best})}_s$.
    \end{algorithmic}
\end{algorithm}

\begin{table*}[htbp]
  \centering
  \caption{Performance comparison results under the cross-platform setting. We plug several debiasing approaches in four evidence-aware fake news detection backbones. Best performances for each backbone are indicated by bold font.}
    \begin{tabular}{cccccc}
    \toprule
    \multirow{2}[2]{*}{Backbone} & \multirow{2}[2]{*}{Debiasing Approach} & \multicolumn{2}{c}{Training: Snopes, Testing: PolitiFact} & \multicolumn{2}{c}{Training: PolitiFact, Testing: Snopes} \\
          &       & F1-Macro & F1-Micro & F1-Macro & F1-Micro \\
    \midrule
    \multirow{6}[2]{*}{BERT} & None  & 0.4932  & 0.5147  & 0.5569  & 0.6204  \\
          & ReW   & 0.5024  & 0.5090  & 0.5555  & 0.6262  \\
          & PoE   & 0.5127  & 0.5065  & 0.5569  & 0.6308  \\
          & CF    & 0.5108  & 0.5093  & 0.5592  & 0.6282  \\
          & EANN  & 0.5167  & 0.5146  & 0.5541  & 0.6311  \\
          & DAL   & \textbf{0.5333} & \textbf{0.5733} & \textbf{0.5654} & \textbf{0.6432} \\
    \midrule
    \multirow{6}[2]{*}{DeClare} & None  & 0.5328  & 0.5350  & 0.5488  & 0.6442  \\
          & ReW   & 0.5366  & 0.5408  & 0.5513  & 0.6468  \\
          & PoE   & 0.5465  & 0.5535  & 0.5562  & 0.6493  \\
          & CF    & 0.5354  & 0.5367  & 0.5509  & 0.6434  \\
          & EANN  & 0.5395  & 0.5422  & 0.5556  & 0.6532  \\
          & DAL   & \textbf{0.5811} & \textbf{0.5813} & \textbf{0.5785} & \textbf{0.6700} \\
    \midrule
    \multirow{6}[2]{*}{MAC} & None  & 0.5484  & 0.5497  & 0.5443  & 0.6320  \\
          & ReW   & 0.5513  & 0.5581  & 0.5478  & 0.6356  \\
          & PoE   & 0.5593  & 0.5648  & 0.5514  & 0.6376  \\
          & CF    & 0.5498  & 0.5548  & 0.5532  & 0.6386  \\
          & EANN  & 0.5564  & 0.5668  & 0.5507  & 0.6406  \\
          & DAL   & \textbf{0.5808} & \textbf{0.5821} & \textbf{0.5787} & \textbf{0.6581} \\
    \midrule
    \multirow{6}[2]{*}{GET} & None  & 0.5473  & 0.5593  & 0.5420  & 0.6581  \\
          & ReW   & 0.5483  & 0.5564  & 0.5503  & 0.6553  \\
          & PoE   & 0.5556  & 0.5648  & 0.5588  & 0.6612  \\
          & CF    & 0.5477  & 0.5674  & 0.5608  & 0.6528  \\
          & EANN  & 0.5543  & 0.5712  & 0.5658  & 0.6616  \\
          & DAL   & \textbf{0.5783} & \textbf{0.5836} & \textbf{0.5805} & \textbf{0.6650} \\
    \bottomrule
    \end{tabular}%
  \label{tab:cross_data}%
\end{table*}%

\subsection{Dual Adversarial Learning}

Inspired by domain-adversarial training \cite{ganin2015unsupervised,ganin2016domain}, which forces models to contain minimum domain information in an adversarial learning manner, we propose to positively optimize the main fake news predictor, while reversely optimize the news-aspect and evidence-aspect debiasing discriminators.
Thus, during the learning of news-evidence reasoning, we can mitigate spurious correlations between news/evidence contents and true/fake news labels.

To avoid trivial solutions, i.e., the reverse optimization of news-aspect and evidence-aspect debiasing discriminators does not affect the parameters in the main fake news predictor ($\theta_f$, $\theta_w$ and $\theta_s$), we need to firstly positively optimize the parameters in the debiasing discriminators ($\theta_n$ and $\theta_e$), while freeze the other parameters.
The news-aspect and evidence-aspect losses respectively are
\begin{equation}
    l_n  = \frac{1}{\left| \mathcal{N} \right|} \sum_{n_i \in \mathcal{N}} {CrossEntropy} \left( { y_i, \hat y^n_i } \right),
\end{equation}
\begin{equation}
    l_e  = \frac{1}{\left| \mathcal{N} \right|} \sum_{n_i \in \mathcal{N}} {CrossEntropy} \left( { y_i, \hat y^e_i } \right),
\end{equation}
and we optimize them as
\begin{equation} \label{eq:op_n}
    \theta^{(k+1)}_n = \mathop {\arg \min }\limits_{\theta_n} l_n \left[ \theta^{(k)}_f,\theta^{(k)}_w,\theta^{(k)}_s,\theta^{(k)}_n \right],
\end{equation}
\begin{equation} \label{eq:op_e}
    \theta^{(k+1)}_e = \mathop {\arg \min }\limits_{\theta_e} l_e \left[ \theta^{(k)}_f,\theta^{(k)}_w,\theta^{(k)}_s,\theta^{(k)}_e \right],
\end{equation}
where $k$ denotes the optimization step.

Then, we freeze $\theta_n$ and $\theta_e$, and optimize the main fake news detector.
The main loss can be calculated as
\begin{equation}
    l_m  = \frac{1}{\left| \mathcal{N} \right|} \sum_{n_i \in \mathcal{N}} {CrossEntropy} \left( { y_i, \hat y_i } \right).
\end{equation}
To mitigate the news and evidence content biases, we construct the overall loss as
\begin{equation}
    l  = l_m - \alpha l_n - \beta l_e,
\end{equation}
where $\alpha$ and $\beta$ are hyper-parameters controlling news-aspect and evidence-aspect debiasing respectively.
In this way, gradients of the debiasing predictors are reversed.
Then, parameters are optimized as
\begin{align} \label{eq:op_m}
\begin{split}
    \theta^{(k+1)}_f,\theta^{(k+1)}_w,\theta^{(k+1)}_s~~~~~~~~~~~~~~~~~~~~~~~~~~~~~~~~~~~~~~~~~~~~~~~~ \\
    ~~~~~~~= \mathop {\arg \min }\limits_{\theta_f,\theta_w,\theta_s} l \left[ \theta^{(k)}_f,\theta^{(k)}_w,\theta^{(k)}_s,\theta^{(k+1)}_n,\theta^{(k+1)}_e \right].
\end{split}
\end{align}

Above two optimization processes are performed alternatively, until convergence.
The dual adversarial debiasing strategy teaches detection models to better perform reasoning and interaction between news and evidences, rather than relying solely on news or evidence contents.
And the whole procedure of DAL is illustrated in Fig. \ref{fig:model} and Alg. \ref{alg:DAL}.

\section{experiments}
\label{sec:experiments}

To evaluate the effectiveness of our proposed DAL approach, we conduct comprehensive experiments under two OOD settings with four state-of-the-art backbones to answer following Research Questions (RQs):
\begin{itemize}
    \item RQ1: Under OOD environments, how well can DAL improve the backbones, and how does DAL perform compared to previous debiasing approaches?
    \item RQ2: How effective are the news-aspect debiasing and evidence-aspect debiasing in DAL?
    \item RQ3: How does DAL perform under different hyperparameter settings?
\end{itemize}
The following subsections describe the details of the experiments, results and analysis.

\subsection{Experimental Configurations}

\subsubsection{Datasets}
We evaluate our DAL on PolitiFact and Snopes datasets which are collected by previous work \cite{hansen2021automatic}.
The news and corresponding labels are from two major fact-checking websites \textbf{PolitiFact}\footnote{https://www.politifact.com/} and \textbf{Snopes}\footnote{https://www.snopes.com/}.
And the evidence are top-10 relevant snippets retrieved by the news.
As we only consider binary classification, we merge \(false\), \(mostly\) \(false\) claims into \(false\) class and the others into \(true\) for Snopes, and likewise merge \(pants\) \(on\) \(fire\), \(false\), \(mostly\) \(false\) into \(false\) claims and the rest to \(true\) for PolitiFact.

\begin{table*}[htbp]
  \centering
  \caption{Performance comparison results under the cross-topic setting. We plug several debiasing approaches in four evidence-aware fake news detection backbones. Best performances for each backbone are indicated by bold font.}
    \begin{tabular}{cccccc}
    \toprule
    \multirow{2}[2]{*}{Backbone} & \multirow{2}[2]{*}{Debiasing Approach} & \multicolumn{2}{c}{Training: PolitiFact, Testing: PolitiFact} & \multicolumn{2}{c}{Training: Snopes, Testing: Snopes} \\
          &       & F1-Macro & F1-Micro & F1-Macro & F1-Micro \\
    \midrule
    \multirow{6}[2]{*}{BERT} & None  & 0.6206  & 0.6206  & 0.5993  & 0.6916  \\
          & ReW   & 0.6223  & 0.6242  & 0.6258  & 0.7143  \\
          & PoE   & 0.6287  & 0.6318  & 0.6512  & 0.7456  \\
          & CF    & 0.6338  & 0.6357  & 0.6374  & 0.7043  \\
          & EANN  & 0.6268  & 0.6281  & 0.6477  & 0.7237  \\
          & DAL   & \textbf{0.6541} & \textbf{0.6566} & \textbf{0.6774} & \textbf{0.7894} \\
    \midrule
    \multirow{6}[2]{*}{DeClare} & None  & 0.6005  & 0.6286  & 0.5941  & 0.6026  \\
          & ReW   & 0.6041  & 0.6296  & 0.5985  & 0.6095  \\
          & PoE   & 0.6125  & 0.6339  & 0.6246  & 0.6458  \\
          & CF    & 0.6153  & 0.6366  & 0.6167  & 0.6387  \\
          & EANN  & 0.6093  & 0.6342  & 0.6281  & 0.6511  \\
          & DAL   & \textbf{0.6451} & \textbf{0.6508} & \textbf{0.6564} & \textbf{0.6977} \\
    \midrule
    \multirow{6}[2]{*}{MAC} & None  & 0.5732  & 0.6117  & 0.6584  & 0.6809  \\
          & ReW   & 0.5844  & 0.6174  & 0.6566  & 0.6767  \\
          & PoE   & 0.6032  & 0.6183  & 0.6582  & 0.6814  \\
          & CF    & 0.5856  & 0.6186  & 0.6541  & 0.6710  \\
          & EANN  & 0.5818  & 0.6203  & 0.6628  & 0.6926  \\
          & DAL   & \textbf{0.6400} & \textbf{0.6405} & \textbf{0.6782} & \textbf{0.7215} \\
    \midrule
    \multirow{6}[2]{*}{GET} & None  & 0.6178  & 0.6358  & 0.6359  & 0.6601  \\
          & ReW   & 0.6242  & 0.6381  & 0.6254  & 0.6565  \\
          & PoE   & 0.6219  & 0.6328  & 0.6335  & 0.6659  \\
          & CF    & 0.6263  & 0.6394  & 0.6278  & 0.6739  \\
          & EANN  & 0.6228  & 0.6388  & 0.6383  & 0.6765  \\
          & DAL   & \textbf{0.6350} & \textbf{0.6441} & \textbf{0.6458} & \textbf{0.7086} \\
    \bottomrule
    \end{tabular}%
  \label{tab:cross_topic}%
\end{table*}%

\subsubsection{Setups}
Following the work\cite{hansen2021automatic, wu2022bias}, We choose F1-Macro and F1-Micro as our evaluation metrics.
To evaluate the effectiveness of our method, we construct OOD settings, including \textbf{cross-platform} and \textbf{cross-topic}.
Under the cross-platform setting, which is in line with the work \cite{hansen2021automatic}, the model is trained and validated on one dataset's training and validation sets respectively, while tested on out-of-dataset testing set (e.g., trained and validated on Snopes while tested on PolitiFact).
Under the cross-topic setting, we first apply the LDA algorithm \cite{blei2003latent} to cluster samples by topics, and split the whole dataset into training, validation, and testing sets to ensure that there is no overlap topics between them.
The training, validation and testing samples are from the same dataset, but share different topics.
It is worth noting that we only tune the hyperparameters on validation sets to ensure that the model cannot access the data distribution of testing sets. And the model early stops when F1-macro does not increase in $10$ epochs. 

\subsubsection{Detection Backbones}
We also choose four state-of-the-art evidence-aware fake news detection model as our backbones:
\begin{itemize}
    \item \textbf{BERT} \cite{kenton2019bert} takes the concatenated news and evidence as input into BERT to model contextual evidence representations.
    \item \textbf{DeClare} \cite{popat2018declare} uses a news specific attention to compute the weights of evidence words and averages the credibility score of each evidence as the final prediction.
    \item \textbf{MAC} \cite{vo2021hierarchical} adopts word-level and document-level attention mechanism to model different granularity interactions between news and evidence.
    \item \textbf{GET} \cite{xu2022evidence} incorporates graph neural network to model textual content, in order to mine fine-grained semantics of news and evidence. 
\end{itemize}

\subsubsection{Debiasing Baselines}
To demonstrate the effectiveness of DAL, we compare it with several debiasing methods for fake news detection:

\begin{itemize}
    \item \textbf{ReW} \cite{schuster2019towards} assigns a low weight for biased samples containing n-grams highly correlated to labels. 
    \item \textbf{PoE} \cite{mahabadi2020end} down-weights the biased samples based on the uneven predictions of the bias-only model.
    \item \textbf{CF} \cite{wu2022bias} adopts counterfactual inference \cite{niu2021counterfactual} to subtract the outputs of a bias-only model to obtain debiased predictions.
    \item \textbf{EANN} \cite{wang2018eann,choi2021using} develops an event adversarial framework to help the model capture event-invariant features in fake news detection.
\end{itemize}

\subsubsection{Implementation Details}
We set the maximum lengths of news and evidences both as $100$, and the optimizer is Adam \cite{kingma2014adam}.
Each news is attached with $10$ pieces of retrieved relevant evidences from websites other than the corresponding news publisher, which are collected in the original datasets \cite{hansen2021automatic}.
We report average results with $5$ different random seeds.
The hyper-parameters $\alpha$ and $\beta$ are both taken from $\{0.001,0.01,0.1,1.0\}$.
Other settings of detection backbones and debiasing baselines follow their original literature.

\begin{table*}[htbp]
  \centering
  \caption{Ablation study under the cross-platform setting. DAL-news and DAL-env indicate DAL with only news- and evidence-aspect debiasing respectively.}
    \begin{tabular}{cccccc}
    \toprule
    \multirow{2}[2]{*}{Backbone} & \multirow{2}[2]{*}{Debiasing Approach} & \multicolumn{2}{c}{Training: Snopes, Testing: PolitiFact} & \multicolumn{2}{c}{Training: PolitiFact, Testing: Snopes} \\
          &       & F1-Macro & F1-Micro & F1-Macro & F1-Micro \\
    \midrule
    \multirow{3}[2]{*}{BERT} & DAL-news & 0.5173  & 0.5671  & 0.5583  & 0.6245  \\
          & DAL-env & 0.5165  & 0.5670  & 0.5520  & 0.6265  \\
          & DAL   & \textbf{0.5333} & \textbf{0.5733} & \textbf{0.5654} & \textbf{0.6432} \\
    \midrule
    \multirow{3}[2]{*}{DeClare} & DAL-news & 0.5715  & 0.5718  & 0.5718  & 0.6456  \\
          & DAL-env & 0.5571  & 0.5615  & 0.5651  & 0.6505  \\
          & DAL   & \textbf{0.5811} & \textbf{0.5813} & \textbf{0.5785} & \textbf{0.6700} \\
    \midrule
    \multirow{3}[2]{*}{MAC} & DAL-news & 0.5694  & 0.5718  & 0.5627  & 0.6482  \\
          & DAL-env & 0.5662  & 0.5689  & 0.5710  & 0.6482  \\
          & DAL   & \textbf{0.5808} & \textbf{0.5821} & \textbf{0.5787} & \textbf{0.6581} \\
    \midrule
    \multirow{3}[2]{*}{GET} & DAL-news & 0.5668  & 0.5678  & 0.5657  & 0.6620  \\
          & DAL-env & 0.5650  & 0.5659  & 0.5703  & 0.6581  \\
          & DAL   & \textbf{0.5783} & \textbf{0.5836} & \textbf{0.5805} & \textbf{0.6650} \\
    \bottomrule
    \end{tabular}%
  \label{tab:ablation_data}%
\end{table*}%

\begin{table*}[htbp]
  \centering
  \caption{Ablation study under the cross-topic setting. DAL-news and DAL-env indicate DAL with only news- and evidence-aspect debiasing respectively.}
    \begin{tabular}{cccccc}
    \toprule
    \multirow{2}[2]{*}{Backbone} & \multirow{2}[2]{*}{Debiasing Approach} & \multicolumn{2}{c}{Training: Snopes, Testing: PolitiFact} & \multicolumn{2}{c}{Training: PolitiFact, Testing: Snopes} \\
          &       & F1-Macro & F1-Micro & F1-Macro & F1-Micro \\
    \midrule
    \multirow{3}[2]{*}{BERT} & DAL-news & 0.6394  & 0.6394  & 0.6423  & 0.7825  \\
          & DAL-env & 0.6256  & 0.6276  & 0.6352  & 0.7537  \\
          & DAL   & \textbf{0.6541} & \textbf{0.6566} & \textbf{0.6774} & \textbf{0.7894} \\
    \midrule
    \multirow{3}[2]{*}{DeClare} & DAL-news & 0.6299  & 0.6397  & 0.6470  & 0.6839  \\
          & DAL-env & 0.6130  & 0.6379  & 0.6415  & 0.6822  \\
          & DAL   & \textbf{0.6451} & \textbf{0.6508} & \textbf{0.6564} & \textbf{0.6977} \\
    \midrule
    \multirow{3}[2]{*}{MAC} & DAL-news & 0.6286  & 0.6353  & 0.6614  & 0.7086  \\
          & DAL-env & 0.6293  & 0.6331  & 0.6622  & 0.7096  \\
          & DAL   & \textbf{0.6400} & \textbf{0.6405} & \textbf{0.6782} & \textbf{0.7215} \\
    \midrule
    \multirow{3}[2]{*}{GET} & DAL-news & 0.6303  & 0.6410  & 0.6364  & 0.6918  \\
          & DAL-env & 0.6329  & 0.6368  & 0.6260  & 0.7017  \\
          & DAL   & \textbf{0.6350} & \textbf{0.6441} & \textbf{0.6458} & \textbf{0.7086} \\
    \bottomrule
    \end{tabular}%
  \label{tab:ablation_topic}%
\end{table*}%

\subsection{Overall Performance (RQ1)}

We compare \themodel with four debiasing methods on four backbones, under cross-platform setting and cross-topic setting respectively.
There is a significant data distribution gap between these two OOD settings, which poses a severe challenge to testing OOD generalization capabilities.
From Tab. \ref{tab:cross_data} and Tab. \ref{tab:cross_topic}, we can have the following observations.

Firstly, under the cross-platform setting, the comparison results are summarized in Tab. \ref{tab:cross_data}.
We can observe that \themodel brings significant improvement for each backbone model on both metrics, compared to the marginal improvement brought by other debiasing competitors.
Notably, our method achieves the top performance consistently, while the performance gains of the other debiasing methods are unstable on different testing sets.
It indicates that the news and evidence content biases in the original datasets can be mitigated to the maximum extent by \themodel effectively and the backbones are facilitated better OOD generalization.
To be more specific, all baselines over DeClare and MAC model can only outperform the backbone without any debiasing method slightly by around $1\%$ measured by F1-Macro and F1-Micro.
In contrast, there is about $3\%$-$4\%$ improvement of DeClare and MAC equipped with DAL measured by each evaluation metric.
Similar phenomena can also be observed on the rest two backbones. 

Secondly, Tab. \ref{tab:cross_topic} summarizes debiasing performance under the cross-topic setting.
The performance of each backbone without any debiasing methods under the cross-topic setting is between that under the in-distribution setting in Tab. \ref{tab:pilot1} and the cross-platform setting in Tab. \ref{tab:cross_data}, indicating that the distribution shift problem exists while is weaker than that in the cross-platform setting.
It is obvious that DAL can still maintain consistent and significant improvement over backbones, compared to its debiasing counterparts.
On the other hand, the performance improvement of baselines varies across different datasets and backbones.
It is worth noting that most baselines other than EANN over the MAC and GET model even underperform the original backbones on the Snopes dataset.

\begin{figure*}
	\centering
	\subfigure[BERT on PolitiFact.]{
		\begin{minipage}[b]{0.23\textwidth}
			\includegraphics[width=1\textwidth]{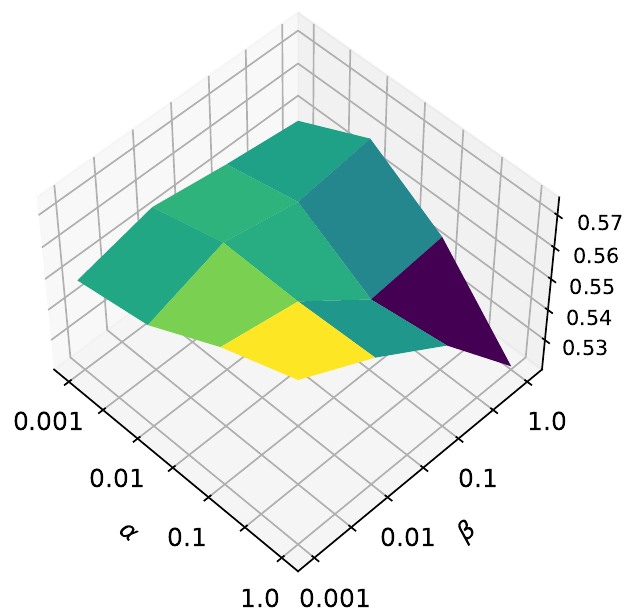}
		\end{minipage}
	}
	\subfigure[BERT on Snopes.]{
		\begin{minipage}[b]{0.23\textwidth}
			\includegraphics[width=1\textwidth]{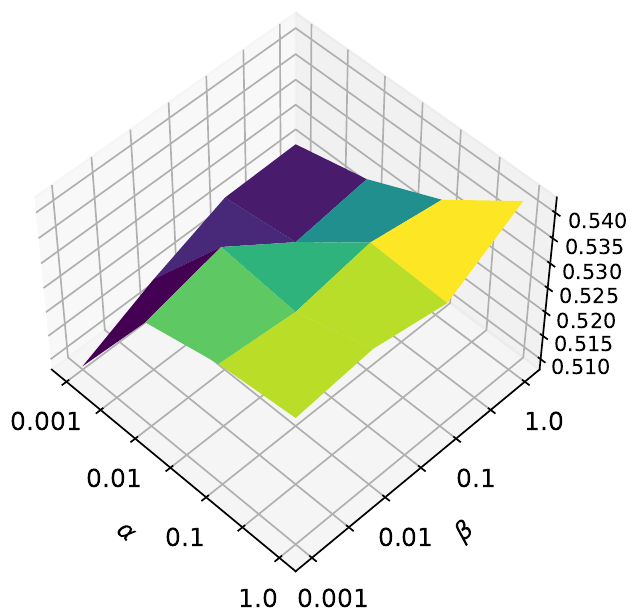}
		\end{minipage}
	}
	\subfigure[DeClare on PolitiFact.]{
		\begin{minipage}[b]{0.23\textwidth}
			\includegraphics[width=1\textwidth]{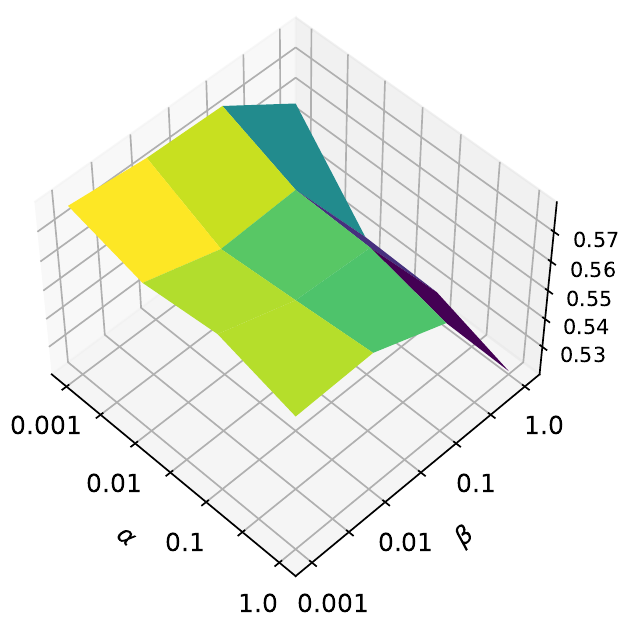}
		\end{minipage}
	}
	\subfigure[DeClare on Snopes.]{
		\begin{minipage}[b]{0.23\textwidth}
			\includegraphics[width=1\textwidth]{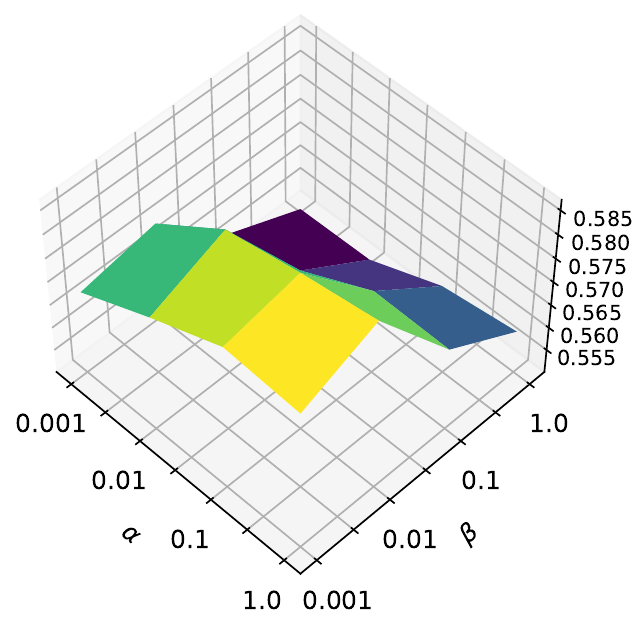}
		\end{minipage}
	}
	\subfigure[MAC on PolitiFact.]{
		\begin{minipage}[b]{0.23\textwidth}
			\includegraphics[width=1\textwidth]{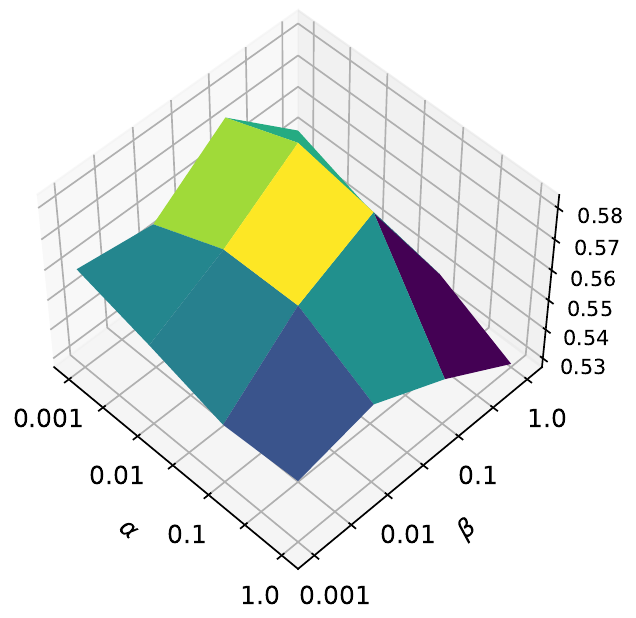}
		\end{minipage}
	}
	\subfigure[MAC on Snopes.]{
		\begin{minipage}[b]{0.23\textwidth}
			\includegraphics[width=1\textwidth]{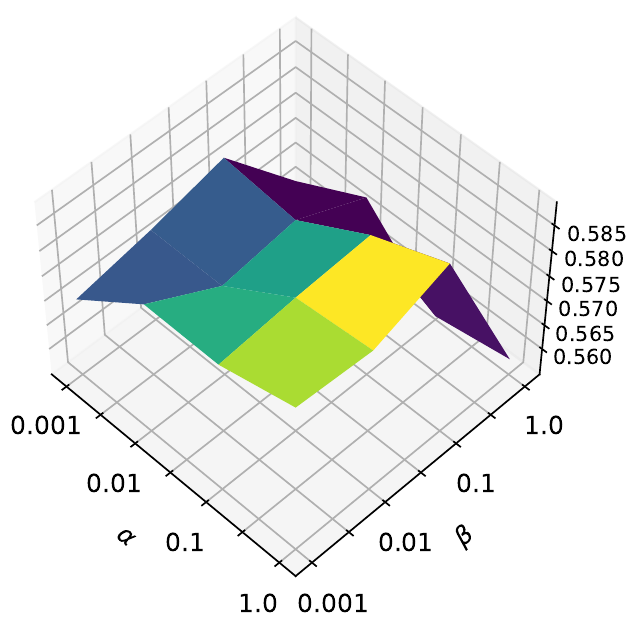}
		\end{minipage}
	}
	\subfigure[GET on PolitiFact.]{
		\begin{minipage}[b]{0.23\textwidth}
			\includegraphics[width=1\textwidth]{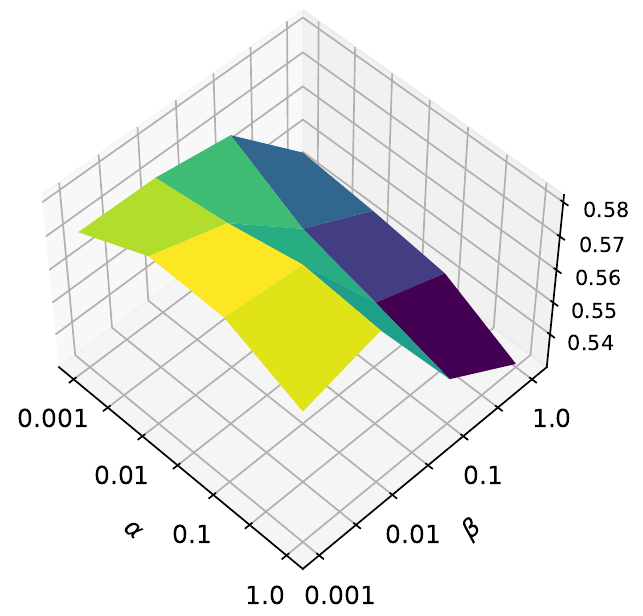}
		\end{minipage}
	}
	\subfigure[GET on Snopes.]{
		\begin{minipage}[b]{0.23\textwidth}
			\includegraphics[width=1\textwidth]{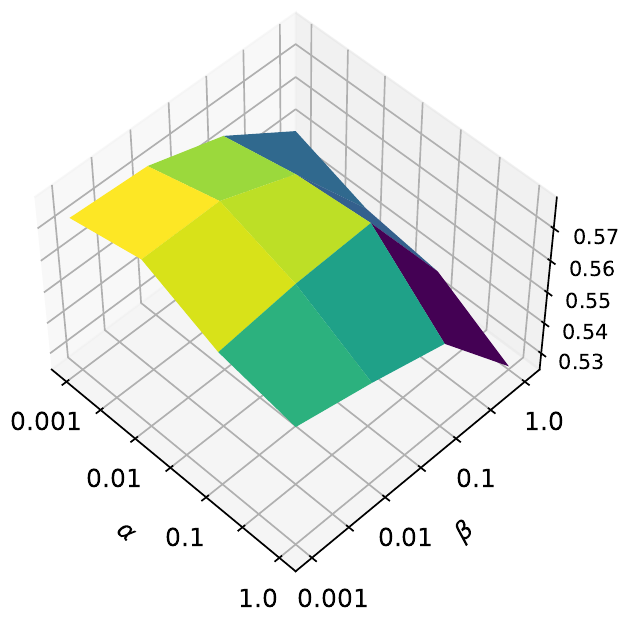}
		\end{minipage}
	}
	\caption{Sensitivity of hyper-parameters, i.e., $\alpha$ and $\beta$, of DAL plugged in four different evidence-aware fake news detection backbones, tested on PoliticFact and Snopes under the cross-platform setting measured by F1-Macro.}
	\label{fig:params_cross}
\end{figure*}

Based on the above two groups of experimental results, we can draw the following conclusions.
(1) Results under the cross-platform setting are much lower than those under the cross-topic setting. This may indicate that the cross-platform setting is with more serious distribution shift issues. Besides the difference in topics, different platforms have other distinguishable factors such as writing styles and audience groups.
(2) Sample weighting approaches bring slight and unstable performance improvements. This shows that the estimation of sample weights is with high variance and hard to be stable, as mentioned in causal inference-related literature. Among them, PoE performs better than ReW. This may indicate that weight estimation based on a sole model is more effective than that based on n-gram statistics.
(3) The CF approach also brings slight performance improvements and even underperforms the original backbones in some cases. This may be due to the excessive bias removal on the training sets leading to harmful effects.
(4) EANN can stably bring improvements over the original backbones. Because it leverages adversarial learning to adaptively mitigate spurious correlations between domain features and labels.
(5) DAL significantly outperforms EANN, though they both incorporate adversarial learning for debiasing. Because EANN, which can be seen as an application of domain-adversarial training, only focuses on common features among different events or topics. In contrast, DAL deeply investigates the specific biases in evidence-aware fake news detection models and accordingly designs suitable adversarial losses. In other words, DAL conducts debiasing in the specific reasoning path in evidence-aware fake news detection, which is the main characteristic of this problem.
(6) DAL achieves the best performances under all the settings. This shows the superiority of mitigating news and evidence content biases with dual adversarial learning.

\subsection{Ablation Study (RQ2)}

To further demonstrate the effects of news-aspect and evidence-aspect debiasing of \themodel, we conduct detailed ablation studies with the four backbones under the two OOD settings and the results are shown in Tab. \ref{tab:ablation_data}.
We denote the DAL variants as follows: (1) \themodel-news: A variant only keeps the news-aspect debiasing. (2) \themodel-env: A variant only keeps the evidence-aspect debiasing. 

According to the results, we can first observe that backbones with \themodel-news and \themodel-env both perform better than the backbones significantly, which indicates the existence of bias from news and evidences and the adversarial debiasing of each bias aspect is beneficial and effective.
What's more, the effectiveness of bias removal from different aspects is similar for most backbones, while DeClare is the exception where \themodel-news outperforms \themodel-env under both settings.
It can be attributed to the different structures of backbones, which determine different bias poisoning from news-aspect and evidence-aspect.

In addition, it is obvious that the proposed \themodel approach outperforms \themodel-news and \themodel-env greatly for all backbones and test settings.
It is reasonable since each aspect of debiasing is effective and demonstrates the importance of mitigating both the impact of news and evidence content biases.
Therefore, dual adversarial debiasing can integrate the effectiveness of each single aspect and further go beyond, making the superiority of \themodel.

\subsection{Sensitivity Analysis (RQ3)}

\begin{figure*}
	\centering
	\subfigure[BERT on PolitiFact.]{
		\begin{minipage}[b]{0.23\textwidth}
			\includegraphics[width=1\textwidth]{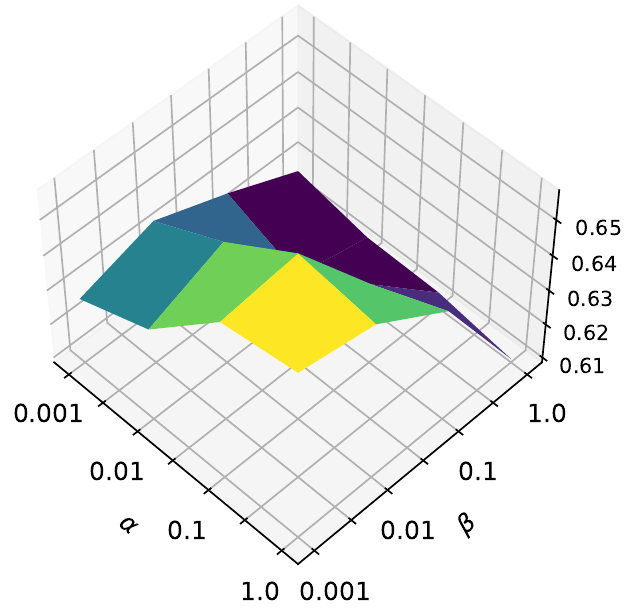}
		\end{minipage}
	}
	\subfigure[BERT on Snopes.]{
		\begin{minipage}[b]{0.23\textwidth}
			\includegraphics[width=1\textwidth]{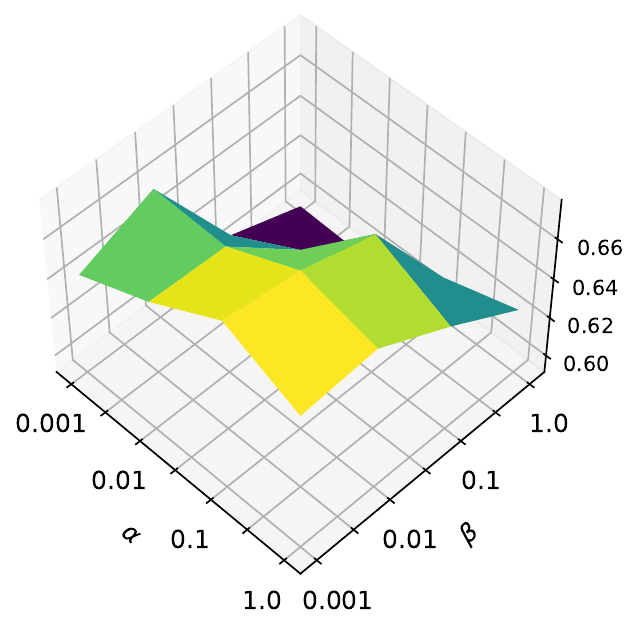}
		\end{minipage}
	}
	\subfigure[DeClare on PolitiFact.]{
		\begin{minipage}[b]{0.23\textwidth}
			\includegraphics[width=1\textwidth]{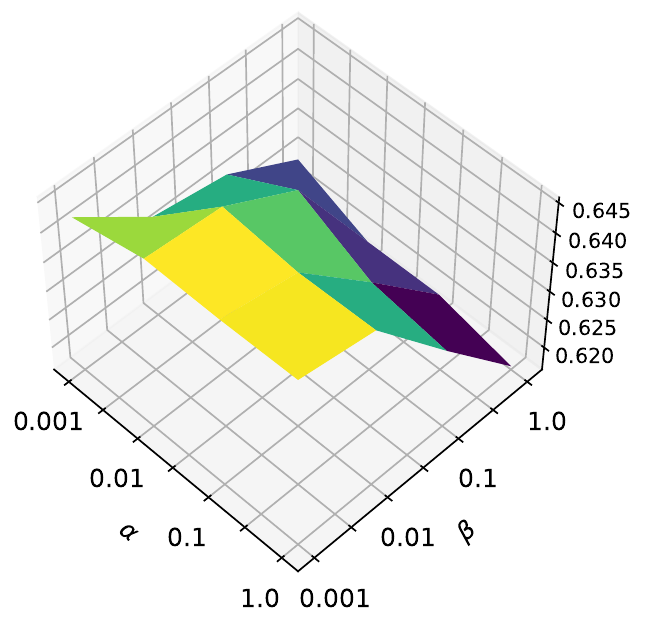}
		\end{minipage}
	}
	\subfigure[DeClare on Snopes.]{
		\begin{minipage}[b]{0.23\textwidth}
			\includegraphics[width=1\textwidth]{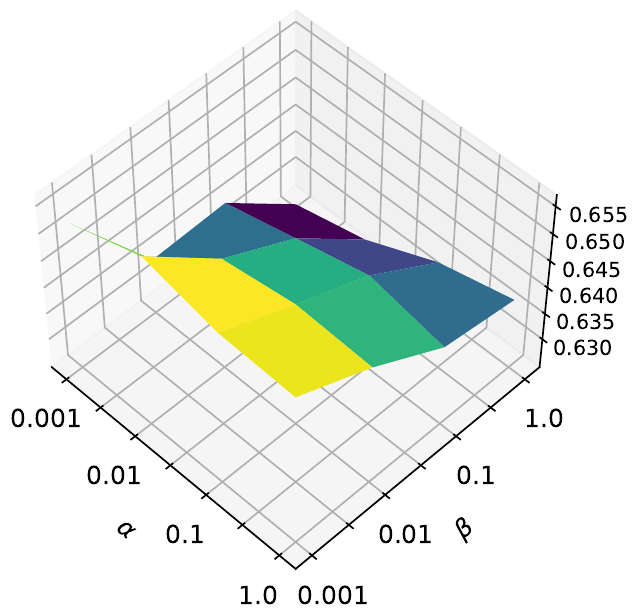}
		\end{minipage}
	}
	\subfigure[MAC on PolitiFact.]{
		\begin{minipage}[b]{0.23\textwidth}
			\includegraphics[width=1\textwidth]{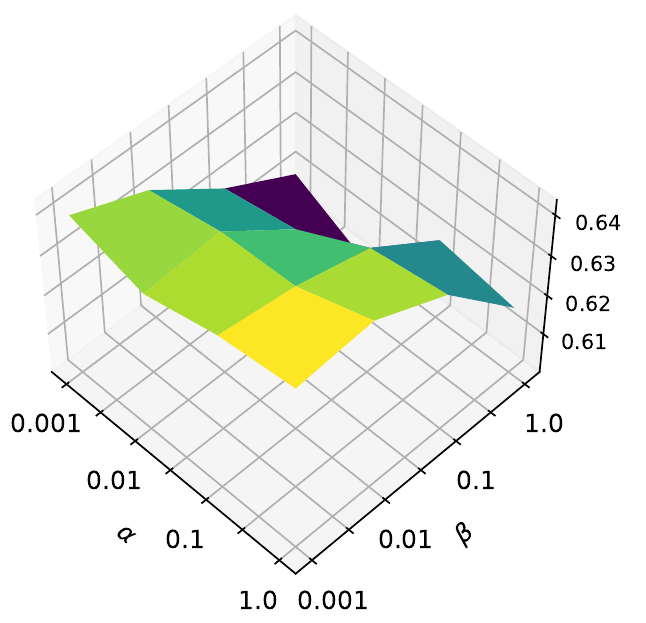}
		\end{minipage}
	}
	\subfigure[MAC on Snopes.]{
		\begin{minipage}[b]{0.23\textwidth}
			\includegraphics[width=1\textwidth]{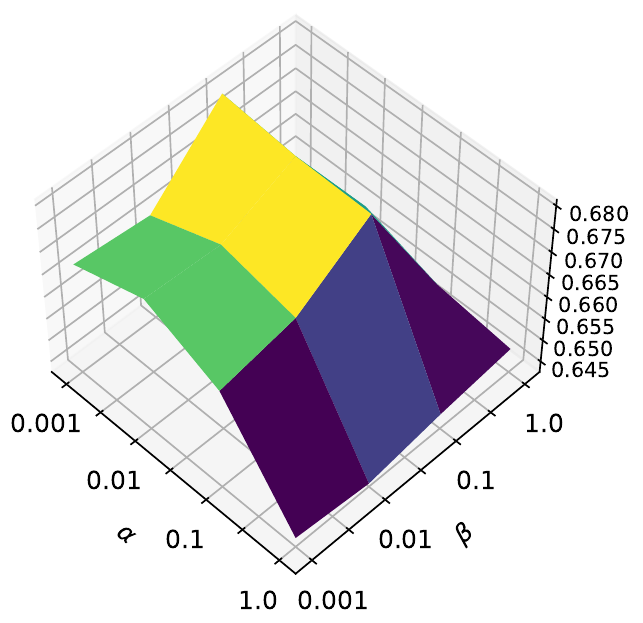}
		\end{minipage}
	}
	\subfigure[GET on PolitiFact.]{
		\begin{minipage}[b]{0.23\textwidth}
			\includegraphics[width=1\textwidth]{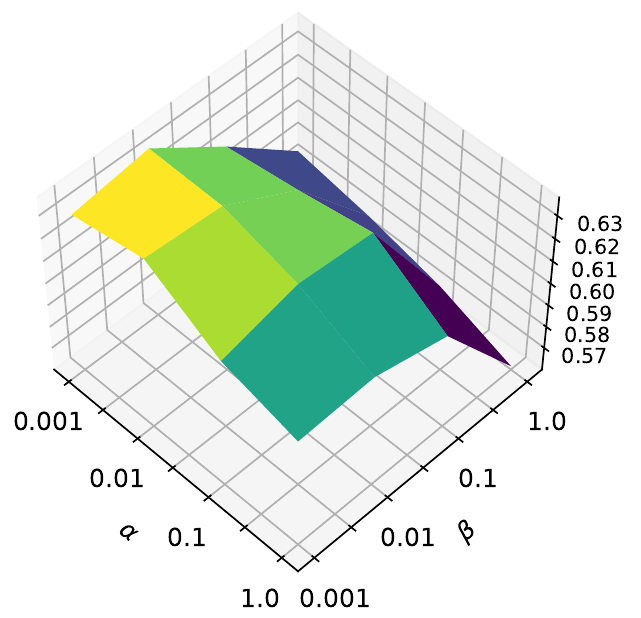}
		\end{minipage}
	}
	\subfigure[GET on Snopes.]{
		\begin{minipage}[b]{0.23\textwidth}
			\includegraphics[width=1\textwidth]{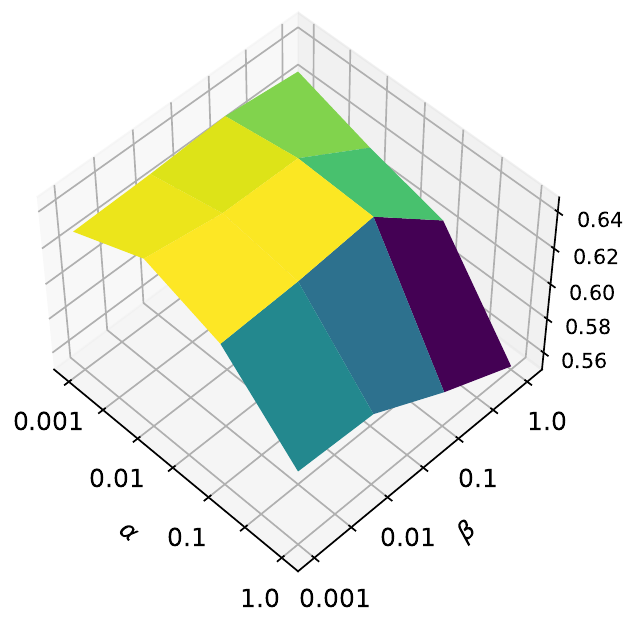}
		\end{minipage}
	}
	\caption{Sensitivity of hyper-parameters, i.e., $\alpha$ and $\beta$, of DAL plugged in four different evidence-aware fake news detection backbones, tested on PoliticFact and Snopes under the cross-topic setting measured by F1-Macro.}
	\label{fig:params_topic}
\end{figure*}

In this section, we test the sensitivity of \themodel to the hyper-parameters $\alpha$ and $\beta$ for four different backbones under the cross-platform and cross-topic settings, with values taken from $\{0.001, 0.01, 0.1, 1.0\}$.
These two hyper-parameters determine the extent of the debiasing of news content bias and evidence content bias, respectively.
The testing results are summarized in Fig. \ref{fig:params_cross} and Fig. \ref{fig:params_topic}.

As shown in Fig. \ref{fig:params_cross}, \themodel shows different sensitivity for different backbones on different datasets.
Taking BERT backbone as an example, the best performance is achieved when $\alpha=1.0$, $\beta=0.001$ on PolitiFact and $\alpha=1.0$, $\beta=1.0$ on Snopes, which indicates that news-aspect debiasing plays an important role for BERT model.
Meanwhile, when $\alpha=0.001$, $\beta=0.001$ on PolitiFact and $\alpha=1.0$, $\beta=0.001$ on Snops, DeClare equipped with DAL has the highest results, which indicates that a slight extent of evidence-aspect debiasing is beneficial for DeClare.
It is mainly due to the bias poison from news and evidence that has distinctions for different backbones, then the optimal hyper-parameters for different backbones are different.
We can also observe similar phenomena under the cross-topic setting, as shown in Fig. \ref{fig:params_topic}.
To be noted, in the rest of our experiments, hyper-parameters are tuned and results are reported according to the best performances on validation sets under different OOD settings.

\section{conclusion}

In this paper, we propose a plug-and-play Dual Adversarial Learning (DAL) approach for debiasing evidence-aware fake news detection models.
DAL incorporates a news-aspect debiasing predictor, and an evidence-aspect debiasing predictor for all the corresponding evidences.
Via reversely optimizing the news-aspect and evidence-aspect discriminators, while positively optimizing the main fake news predictor, DAL can mitigate the spurious correlations between news/evidence contents and true/fake news labels.
Experiments under two different OOD settings and with four evidence-aware backbones, strongly demonstrate the effectiveness and stability of the DAL approach.

\ifCLASSOPTIONcompsoc
  \section*{Acknowledgments}
\else
  \section*{Acknowledgment}
\fi

The authors would like to thank the anonymous reviewers for their valuable comments and suggestions allowing them to improve the quality of this paper. 
This work is jointly sponsored by National Natural Science Foundation of China (62206291, 62141608, and 62236010).



\bibliographystyle{IEEEtran}
\bibliography{tkde}

\begin{thebibliography}{10}
\providecommand{\url}[1]{#1}
\csname url@samestyle\endcsname
\providecommand{\newblock}{\relax}
\providecommand{\bibinfo}[2]{#2}
\providecommand{\BIBentrySTDinterwordspacing}{\spaceskip=0pt\relax}
\providecommand{\BIBentryALTinterwordstretchfactor}{4}
\providecommand{\BIBentryALTinterwordspacing}{\spaceskip=\fontdimen2\font plus
\BIBentryALTinterwordstretchfactor\fontdimen3\font minus
  \fontdimen4\font\relax}
\providecommand{\BIBforeignlanguage}[2]{{%
\expandafter\ifx\csname l@#1\endcsname\relax
\typeout{** WARNING: IEEEtran.bst: No hyphenation pattern has been}%
\typeout{** loaded for the language `#1'. Using the pattern for}%
\typeout{** the default language instead.}%
\else
\language=\csname l@#1\endcsname
\fi
#2}}
\providecommand{\BIBdecl}{\relax}
\BIBdecl

\bibitem{Naeem2020TheC}
S.~B. Naeem and R.~Bhatti, ``The covid-19 ‘infodemic’: a new front for
  information professionals,'' \emph{Health Information and Libraries Journal},
  2020.

\bibitem{castillo2011information}
C.~Castillo, M.~Mendoza, and B.~Poblete, ``Information credibility on
  twitter,'' in \emph{Proceedings of the International Conference on World Wide
  Web}, 2011, pp. 675--684.

\bibitem{volkova2017separating}
S.~Volkova, K.~Shaffer, J.~Y. Jang, and N.~Hodas, ``Separating facts from
  fiction: Linguistic models to classify suspicious and trusted news posts on
  twitter,'' in \emph{Proceedings of the Annual Meeting of the Association for
  Computational Linguistics}, 2017, pp. 647--653.

\bibitem{giachanou2019leveraging}
A.~Giachanou, P.~Rosso, and F.~Crestani, ``Leveraging emotional signals for
  credibility detection,'' in \emph{Proceedings of the International ACM SIGIR
  Conference on Research and Development in Information Retrieval}, 2019, pp.
  877--880.

\bibitem{przybyla2020capturing}
P.~Przybyla, ``Capturing the style of fake news,'' in \emph{Proceedings of the
  AAAI Conference on Artificial Intelligence}, 2020, pp. 490--497.

\bibitem{jin2017multimodal}
Z.~Jin, J.~Cao, H.~Guo, Y.~Zhang, and J.~Luo, ``Multimodal fusion with
  recurrent neural networks for rumor detection on microblogs,'' in
  \emph{Proceedings of the ACM International Conference on Multimedia}, 2017,
  pp. 795--816.

\bibitem{khattar2019mvae}
D.~Khattar, J.~S. Goud, M.~Gupta, and V.~Varma, ``Mvae: Multimodal variational
  autoencoder for fake news detection,'' in \emph{The World Wide Web
  Conference}, 2019, pp. 2915--2921.

\bibitem{qian2021hierarchical}
S.~Qian, J.~Wang, J.~Hu, Q.~Fang, and C.~Xu, ``Hierarchical multi-modal
  contextual attention network for fake news detection,'' in \emph{Proceedings
  of the International ACM SIGIR Conference on Research and Development in
  Information Retrieval}, 2021, pp. 153--162.

\bibitem{kwon2013prominent}
S.~Kwon, M.~Cha, K.~Jung, W.~Chen, and Y.~Wang, ``Prominent features of rumor
  propagation in online social media,'' in \emph{IEEE International Conference
  on Data Mining}, 2013, pp. 1103--1108.

\bibitem{ma2015detect}
J.~Ma, W.~Gao, Z.~Wei, Y.~Lu, and K.-F. Wong, ``Detect rumors using time series
  of social context information on microblogging websites,'' in
  \emph{Proceedings of the ACM International on Conference on Information and
  Knowledge Management}, 2015, pp. 1751--1754.

\bibitem{ma2016detecting}
J.~Ma, W.~Gao, P.~Mitra, S.~Kwon, B.~J. Jansen, K.-F. Wong, and M.~Cha,
  ``Detecting rumors from microblogs with recurrent neural networks,'' in
  \emph{Proceedings of the International Joint Conference on Artificial
  Intelligence}, 2016, pp. 3818--3824.

\bibitem{bian2020rumor}
T.~Bian, X.~Xiao, T.~Xu, P.~Zhao, W.~Huang, Y.~Rong, and J.~Huang, ``Rumor
  detection on social media with bi-directional graph convolutional networks,''
  in \emph{Proceedings of the AAAI Conference on Artificial Intelligence},
  2020, pp. 549--556.

\bibitem{popat2017truth}
K.~Popat, S.~Mukherjee, J.~Str{\"o}tgen, and G.~Weikum, ``Where the truth lies:
  Explaining the credibility of emerging claims on the web and social media,''
  in \emph{Proceedings of the International Conference on World Wide Web
  Companion}, 2017, pp. 1003--1012.

\bibitem{popat2018declare}
K.~Popat, S.~Mukherjee, A.~Yates, and G.~Weikum, ``Declare: Debunking fake news
  and false claims using evidence-aware deep learning,'' in \emph{Proceedings
  of the Conference on Empirical Methods in Natural Language Processing}, 2018,
  pp. 22--32.

\bibitem{vo2021hierarchical}
N.~Vo and K.~Lee, ``Hierarchical multi-head attentive network for
  evidence-aware fake news detection,'' in \emph{Proceedings of the Conference
  of the European Chapter of the Association for Computational Linguistics},
  2021, pp. 965--975.

\bibitem{xu2022evidence}
W.~Xu, J.~Wu, Q.~Liu, S.~Wu, and L.~Wang, ``Evidence-aware fake news detection
  with graph neural networks,'' in \emph{Proceedings of the Web Conference},
  2022, pp. 2501--2510.

\bibitem{hansen2021automatic}
C.~Hansen, C.~Hansen, and L.~C. Lima, ``Automatic fake news detection: Are
  models learning to reason?'' in \emph{Proceedings of the Annual Meeting of
  the Association for Computational Linguistics}, 2021, pp. 80--86.

\bibitem{lin2022detect}
H.~Lin, J.~Ma, L.~Chen, Z.~Yang, M.~Cheng, and G.~Chen, ``Detect rumors in
  microblog posts for low-resource domains via adversarial contrastive
  learning,'' in \emph{Proceedings of the Conference of the North American
  Chapter of the Association for Computational Linguistics}, 2022.

\bibitem{wang2018eann}
Y.~Wang, F.~Ma, Z.~Jin, Y.~Yuan, G.~Xun, K.~Jha, L.~Su, and J.~Gao, ``Eann:
  Event adversarial neural networks for multi-modal fake news detection,'' in
  \emph{Proceedings of the ACM SIGKDD international conference on knowledge
  discovery \& data mining}, 2018, pp. 849--857.

\bibitem{choi2021using}
H.~Choi and Y.~Ko, ``Using topic modeling and adversarial neural networks for
  fake news video detection,'' in \emph{Proceedings of the ACM International
  Conference on Information \& Knowledge Management}, 2021, pp. 2950--2954.

\bibitem{nan2021mdfend}
Q.~Nan, J.~Cao, Y.~Zhu, Y.~Wang, and J.~Li, ``Mdfend: Multi-domain fake news
  detection,'' in \emph{Proceedings of the ACM International Conference on
  Information \& Knowledge Management}, 2021, pp. 3343--3347.

\bibitem{zhu2022memory}
Y.~Zhu, Q.~Sheng, J.~Cao, Q.~Nan, K.~Shu, M.~Wu, J.~Wang, and F.~Zhuang,
  ``Memory-guided multi-view multi-domain fake news detection,'' \emph{IEEE
  Transactions on Knowledge and Data Engineering}, 2022.

\bibitem{silva2021embracing}
A.~Silva, L.~Luo, S.~Karunasekera, and C.~Leckie, ``Embracing domain
  differences in fake news: Cross-domain fake news detection using multi-modal
  data,'' in \emph{Proceedings of the AAAI Conference on Artificial
  Intelligence}, 2021, pp. 557--565.

\bibitem{nan2022improving}
Q.~Nan, D.~Wang, Y.~Zhu, Q.~Sheng, Y.~Shi, J.~Cao, and J.~Li, ``Improving fake
  news detection of influential domain via domain-and instance-level
  transfer,'' in \emph{Proceedings of the International Conference on
  Computational Linguistics}, 2022, pp. 2834--2848.

\bibitem{niu2021counterfactual}
Y.~Niu, K.~Tang, H.~Zhang, Z.~Lu, X.-S. Hua, and J.-R. Wen, ``Counterfactual
  vqa: A cause-effect look at language bias,'' in \emph{Proceedings of the
  IEEE/CVF Conference on Computer Vision and Pattern Recognition}, 2021, pp.
  12\,700--12\,710.

\bibitem{wu2022bias}
J.~Wu, Q.~Liu, W.~Xu, and S.~Wu, ``Bias mitigation for evidence-aware fake news
  detection by causal intervention,'' in \emph{Proceedings of the International
  ACM SIGIR Conference on Research and Development in Information Retrieval},
  2022, pp. 2308--2313.

\bibitem{thorne2018fever}
J.~Thorne, A.~Vlachos, C.~Christodoulopoulos, and A.~Mittal, ``Fever: a
  large-scale dataset for fact extraction and verification,'' in
  \emph{Proceedings of the Conference of the North American Chapter of the
  Association for Computational Linguistics}, 2018, pp. 809--819.

\bibitem{zhou2019gear}
J.~Zhou, X.~Han, C.~Yang, Z.~Liu, L.~Wang, C.~Li, and M.~Sun, ``Gear:
  Graph-based evidence aggregating and reasoning for fact verification,'' in
  \emph{Proceedings of the Annual Meeting of the Association for Computational
  Linguistics}, 2019, pp. 892--901.

\bibitem{liu2020fine}
Z.~Liu, C.~Xiong, M.~Sun, and Z.~Liu, ``Fine-grained fact verification with
  kernel graph attention network,'' in \emph{Proceedings of the Annual Meeting
  of the Association for Computational Linguistics}, 2020, pp. 7342--7351.

\bibitem{lee2021crossaug}
M.~Lee, S.~Won, J.~Kim, H.~Lee, C.~Park, and K.~Jung, ``Crossaug: A contrastive
  data augmentation method for debiasing fact verification models,'' in
  \emph{Proceedings of the ACM International Conference on Information \&
  Knowledge Management}, 2021, pp. 3181--3185.

\bibitem{schuster2019towards}
T.~Schuster, D.~Shah, Y.~J.~S. Yeo, D.~R.~F. Ortiz, E.~Santus, and R.~Barzilay,
  ``Towards debiasing fact verification models,'' in \emph{Proceedings of the
  Conference on Empirical Methods in Natural Language Processing}, 2019, pp.
  3419--3425.

\bibitem{ganin2015unsupervised}
Y.~Ganin and V.~Lempitsky, ``Unsupervised domain adaptation by
  backpropagation,'' in \emph{International Conference on Machine Learning},
  2015, pp. 1180--1189.

\bibitem{ganin2016domain}
Y.~Ganin, E.~Ustinova, H.~Ajakan, P.~Germain, H.~Larochelle, F.~Laviolette,
  M.~Marchand, and V.~Lempitsky, ``Domain-adversarial training of neural
  networks,'' \emph{The Journal of Machine Learning Research}, vol.~17, no.~1,
  pp. 2096--2030, 2016.

\bibitem{rashkin2017truth}
H.~Rashkin, E.~Choi, J.~Y. Jang, S.~Volkova, and Y.~Choi, ``Truth of varying
  shades: Analyzing language in fake news and political fact-checking,'' in
  \emph{Proceedings of the Conference on Empirical Methods in Natural Language
  Processing}, 2017, pp. 2931--2937.

\bibitem{zhang2021mining}
X.~Zhang, J.~Cao, X.~Li, Q.~Sheng, L.~Zhong, and K.~Shu, ``Mining dual emotion
  for fake news detection,'' in \emph{Proceedings of the Web Conference}, 2021,
  pp. 3465--3476.

\bibitem{kenton2019bert}
J.~D. M.-W.~C. Kenton and L.~K. Toutanova, ``Bert: Pre-training of deep
  bidirectional transformers for language understanding,'' in \emph{Proceedings
  of the Annual Conference of the North American Chapter of the Association for
  Computational Linguistics}, 2019, pp. 4171--4186.

\bibitem{kaliyar2021fakebert}
R.~K. Kaliyar, A.~Goswami, and P.~Narang, ``Fakebert: Fake news detection in
  social media with a bert-based deep learning approach,'' \emph{Multimedia
  Tools and Applications}, vol.~80, no.~8, pp. 11\,765--11\,788, 2021.

\bibitem{sheng2022zoom}
Q.~Sheng, J.~Cao, X.~Zhang, R.~Li, D.~Wang, and Y.~Zhu, ``Zoom out and observe:
  News environment perception for fake news detection,'' in \emph{Proceedings
  of the Annual Meeting of the Association for Computational Linguistics},
  2022, pp. 4543--4556.

\bibitem{jin2016novel}
Z.~Jin, J.~Cao, Y.~Zhang, J.~Zhou, and Q.~Tian, ``Novel visual and statistical
  image features for microblogs news verification,'' \emph{IEEE Transactions on
  Multimedia}, vol.~19, no.~3, pp. 598--608, 2016.

\bibitem{tan2020detecting}
R.~Tan, B.~Plummer, and K.~Saenko, ``Detecting cross-modal inconsistency to
  defend against neural fake news,'' in \emph{Proceedings of the Conference on
  Empirical Methods in Natural Language Processing}, 2020, pp. 2081--2106.

\bibitem{qi2021improving}
P.~Qi, J.~Cao, X.~Li, H.~Liu, Q.~Sheng, X.~Mi, Q.~He, Y.~Lv, C.~Guo, and Y.~Yu,
  ``Improving fake news detection by using an entity-enhanced framework to fuse
  diverse multimodal clues,'' in \emph{Proceedings of the ACM International
  Conference on Multimedia}, 2021, pp. 1212--1220.

\bibitem{chen2022cross}
Y.~Chen, D.~Li, P.~Zhang, J.~Sui, Q.~Lv, L.~Tun, and L.~Shang, ``Cross-modal
  ambiguity learning for multimodal fake news detection,'' in \emph{Proceedings
  of the Web Conference}, 2022, pp. 2897--2905.

\bibitem{ma2018rumor}
J.~Ma, W.~Gao, and K.-F. Wong, ``Rumor detection on twitter with
  tree-structured recursive neural networks,'' in \emph{Proceedings of the
  Annual Meeting of the Association for Computational Linguistics=}, 2018, pp.
  1980--1989.

\bibitem{yu2017convolutional}
F.~Yu, Q.~Liu, S.~Wu, L.~Wang, and T.~Tan, ``A convolutional approach for
  misinformation identification,'' in \emph{Proceedings of the International
  Joint Conference on Artificial Intelligence}, 2017, pp. 3901--3907.

\bibitem{liu2018early}
Y.~Liu and Y.-F. Wu, ``Early detection of fake news on social media through
  propagation path classification with recurrent and convolutional networks,''
  in \emph{Proceedings of the AAAI Conference on Artificial Intelligence},
  2018.

\bibitem{liu2018mining}
Q.~Liu, F.~Yu, S.~Wu, and L.~Wang, ``Mining significant microblogs for
  misinformation identification: an attention-based approach,'' \emph{ACM
  Transactions on Intelligent Systems and Technology (TIST)}, vol.~9, no.~5,
  pp. 1--20, 2018.

\bibitem{ma2019detect}
J.~Ma, W.~Gao, and K.-F. Wong, ``Detect rumors on twitter by promoting
  information campaigns with generative adversarial learning,'' in \emph{The
  World Wide Web Conference}, 2019, pp. 3049--3055.

\bibitem{kipf2017semi}
T.~N. Kipf and M.~Welling, ``Semi-supervised classification with graph
  convolutional networks,'' in \emph{International Conference on Learning
  Representations}, 2017.

\bibitem{lu2020gcan}
Y.-J. Lu and C.-T. Li, ``Gcan: Graph-aware co-attention networks for
  explainable fake news detection on social media,'' in \emph{Proceedings of
  the Annual Meeting of the Association for Computational Linguistics}, 2020,
  pp. 505--514.

\bibitem{nguyen2020fang}
V.-H. Nguyen, K.~Sugiyama, P.~Nakov, and M.-Y. Kan, ``Fang: Leveraging social
  context for fake news detection using graph representation,'' in
  \emph{Proceedings of the ACM International Conference on Information \&
  Knowledge Management}, 2020, pp. 1165--1174.

\bibitem{sun2022rumor}
T.~Sun, Z.~Qian, S.~Dong, P.~Li, and Q.~Zhu, ``Rumor detection on social media
  with graph adversarial contrastive learning,'' in \emph{Proceedings of the
  Web Conference}, 2022, pp. 2789--2797.

\bibitem{sun2022structure}
X.~Sun, H.~Yin, B.~Liu, Q.~Meng, J.~Cao, A.~Zhou, and H.~Chen, ``Structure
  learning via meta-hyperedge for dynamic rumor detection,'' \emph{IEEE
  Transactions on Knowledge and Data Engineering}, 2022.

\bibitem{jin2022towards}
Y.~Jin, X.~Wang, R.~Yang, Y.~Sun, W.~Wang, H.~Liao, and X.~Xie, ``Towards
  fine-grained reasoning for fake news detection,'' in \emph{Proceedings of the
  AAAI Conference on Artificial Intelligence}, 2022, pp. 5746--5754.

\bibitem{yang2022reinforcement}
R.~Yang, X.~Wang, Y.~Jin, C.~Li, J.~Lian, and X.~Xie, ``Reinforcement subgraph
  reasoning for fake news detection,'' in \emph{Proceedings of the ACM SIGKDD
  Conference on Knowledge Discovery and Data Mining}, 2022, pp. 2253--2262.

\bibitem{sun2022ddgcn}
M.~Sun, X.~Zhang, J.~Zheng, and G.~Ma, ``Ddgcn: Dual dynamic graph
  convolutional networks for rumor detection on social media,'' in
  \emph{Proceedings of the AAAI Conference on Artificial Intelligence}, 2022,
  pp. 4611--4619.

\bibitem{ma2019sentence}
J.~Ma, W.~Gao, S.~Joty, and K.-F. Wong, ``Sentence-level evidence embedding for
  claim verification with hierarchical attention networks,'' in
  \emph{Proceedings of the Annual Meeting of the Association for Computational
  Linguistics}, 2019, pp. 2561--2571.

\bibitem{wu2021evidenceIJCAI}
L.~Wu, Y.~Rao, X.~Yang, W.~Wang, and A.~Nazir, ``Evidence-aware hierarchical
  interactive attention networks for explainable claim verification,'' in
  \emph{Proceedings of the International Conference on International Joint
  Conferences on Artificial Intelligence}, 2021, pp. 1388--1394.

\bibitem{wu2021evidenceAAAI}
L.~Wu, Y.~Rao, L.~Sun, and W.~He, ``Evidence inference networks for
  interpretable claim verification,'' in \emph{Proceedings of the AAAI
  Conference on Artificial Intelligence}, 2021, pp. 14\,058--14\,066.

\bibitem{wu2021unified}
L.~Wu, Y.~Rao, Y.~Lan, L.~Sun, and Z.~Qi, ``Unified dual-view cognitive model
  for interpretable claim verification,'' in \emph{Proceedings of the Annual
  Meeting of the Association for Computational Linguistics}, 2021, pp. 59--68.

\bibitem{wu2022adversarial}
J.~Wu, W.~Xu, Q.~Liu, S.~Wu, and L.~Wang, ``Adversarial contrastive learning
  for evidence-aware fake news detection with graph neural networks,''
  \emph{arXiv preprint arXiv:2210.05498}, 2022.

\bibitem{sheng2021integrating}
Q.~Sheng, X.~Zhang, J.~Cao, and L.~Zhong, ``Integrating pattern-and fact-based
  fake news detection via model preference learning,'' in \emph{Proceedings of
  the ACM International Conference on Information \& Knowledge Management},
  2021, pp. 1640--1650.

\bibitem{schnabel2016recommendations}
T.~Schnabel, A.~Swaminathan, A.~Singh, N.~Chandak, and T.~Joachims,
  ``Recommendations as treatments: Debiasing learning and evaluation,'' in
  \emph{ICML}, 2016, pp. 1670--1679.

\bibitem{zhang2021causal}
Y.~Zhang, F.~Feng, X.~He, T.~Wei, C.~Song, G.~Ling, and Y.~Zhang, ``Causal
  intervention for leveraging popularity bias in recommendation,'' in
  \emph{Proceedings of the International ACM SIGIR conference on Research and
  Development in Information Retrieval}, 2021.

\bibitem{hu2022causal}
L.~Hu, Z.~Chen, Z.~Z.~J. Yin, and L.~Nie, ``Causal inference for leveraging
  image-text matching bias in multi-modal fake news detection,'' \emph{IEEE
  Transactions on Knowledge and Data Engineering}, 2022.

\bibitem{clark2019don}
C.~Clark, M.~Yatskar, and L.~Zettlemoyer, ``Don’t take the easy way out:
  Ensemble based methods for avoiding known dataset biases,'' in
  \emph{Proceedings of the Conference on Empirical Methods in Natural Language
  Processing}, 2019, pp. 4069--4082.

\bibitem{mahabadi2020end}
R.~K. Mahabadi, Y.~Belinkov, and J.~Henderson, ``End-to-end bias mitigation by
  modelling biases in corpora,'' in \emph{Proceedings of the Annual Meeting of
  the Association for Computational Linguistics}, 2020, pp. 8706--8716.

\bibitem{zhu2022generalizing}
Y.~Zhu, Q.~Sheng, J.~Cao, S.~Li, D.~Wang, and F.~Zhuang, ``Generalizing to the
  future: Mitigating entity bias in fake news detection,'' in \emph{Proceedings
  of the International ACM SIGIR Conference on Research and Development in
  Information Retrieval}, 2022, pp. 2120--2125.

\bibitem{xu2023counterfactual}
W.~Xu, Q.~Liu, S.~Wu, and L.~Wang, ``Counterfactual debiasing for fact
  verification,'' in \emph{Proceedings of the Annual Conference of the North
  American Chapter of the Association for Computational Linguistics}, 2023, pp.
  6777--6789.

\bibitem{wei2019eda}
J.~Wei and K.~Zou, ``Eda: Easy data augmentation techniques for boosting
  performance on text classification tasks,'' in \emph{Proceedings of the
  Conference on Empirical Methods in Natural Language Processing}, 2019, pp.
  6382--6388.

\bibitem{lin2023zero}
H.~Lin, P.~Yi, J.~Ma, H.~Jiang, Z.~Luo, S.~Shi, and R.~Liu, ``Zero-shot rumor
  detection with propagation structure via prompt learning,'' in
  \emph{Proceedings of the AAAI Conference on Artificial Intelligence}, 2023.

\bibitem{dey2017gate}
R.~Dey and F.~M. Salem, ``Gate-variants of gated recurrent unit (gru) neural
  networks,'' in \emph{IEEE International Midwest Symposium on Circuits and
  Systems}, 2017, pp. 1597--1600.

\bibitem{huang2015bidirectional}
Z.~Huang, W.~Xu, and K.~Yu, ``Bidirectional lstm-crf models for sequence
  tagging,'' \emph{arXiv preprint arXiv:1508.01991}, 2015.

\bibitem{kipf2016semi}
T.~N. Kipf and M.~Welling, ``Semi-supervised classification with graph
  convolutional networks,'' \emph{arXiv preprint arXiv:1609.02907}, 2016.

\bibitem{blei2003latent}
D.~M. Blei, A.~Y. Ng, and M.~I. Jordan, ``Latent dirichlet allocation,''
  \emph{Journal of machine Learning research}, vol.~3, no. Jan, pp. 993--1022,
  2003.

\bibitem{kingma2014adam}
D.~P. Kingma and J.~Ba, ``Adam: A method for stochastic optimization,''
  \emph{arXiv preprint arXiv:1412.6980}, 2014.

\end{thebibliography}
%



%

\begin{IEEEbiography}[{\includegraphics[width=1in,height=1.25in,keepaspectratio]{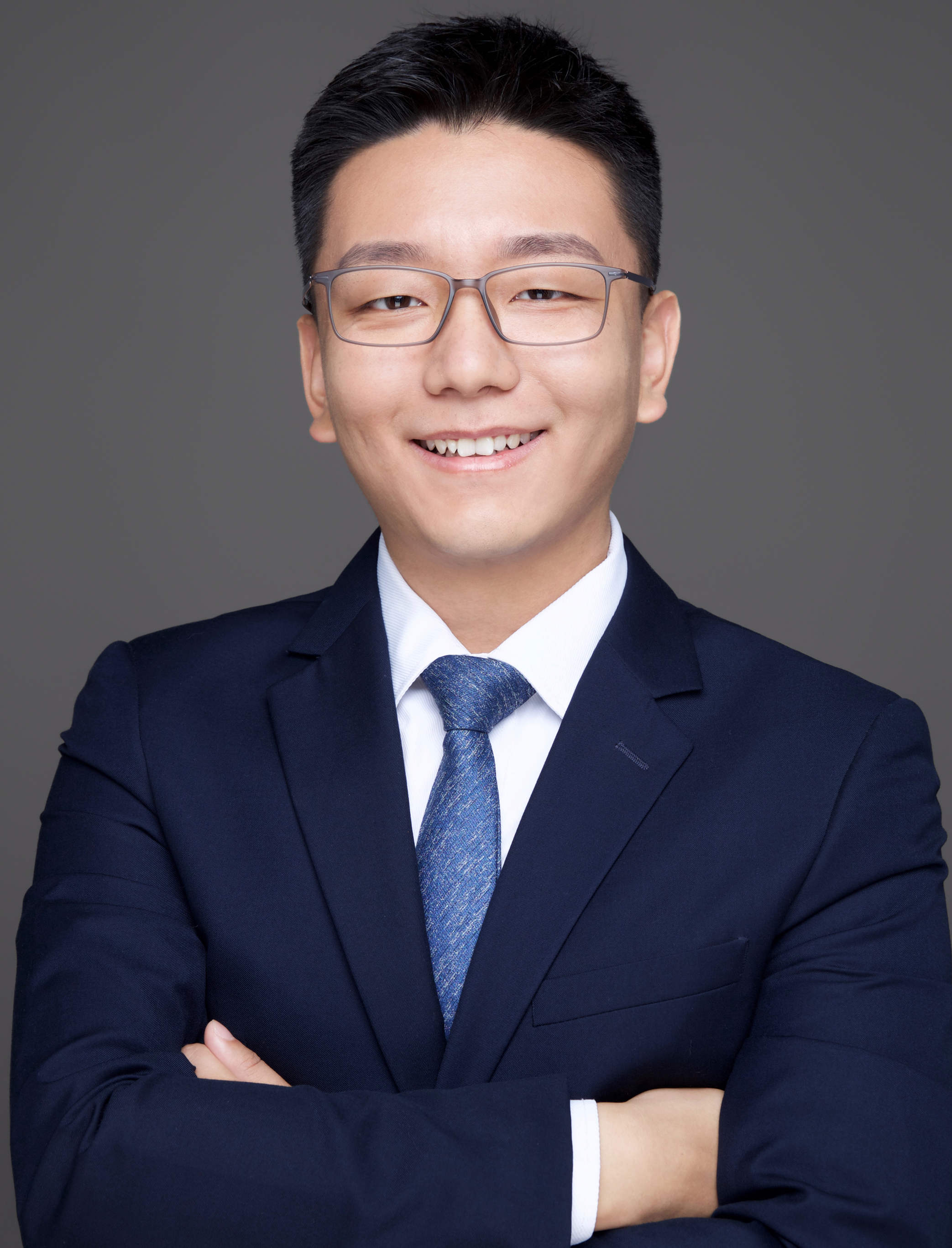}}]{Qiang Liu}
is an Associate Professor with the Center for Research on Intelligent Perception and Computing (CRIPAC), State Key Laboratory of Multimodal Artificial Intelligence Systems (MAIS), Institute of Automation, Chinese Academy of Sciences (CASIA). He received his PhD degree from CASIA. Currently, his research interests include data mining, misinformation detection, LLM safety and AI for science. He has published papers in top-tier journals and conferences, such as IEEE TKDE, AAAI, NeurIPS, KDD, WWW, SIGIR, CIKM, ICDM, ACL and EMNLP.
\end{IEEEbiography}

\begin{IEEEbiography}[{\includegraphics[width=1in,height=1.25in,keepaspectratio]{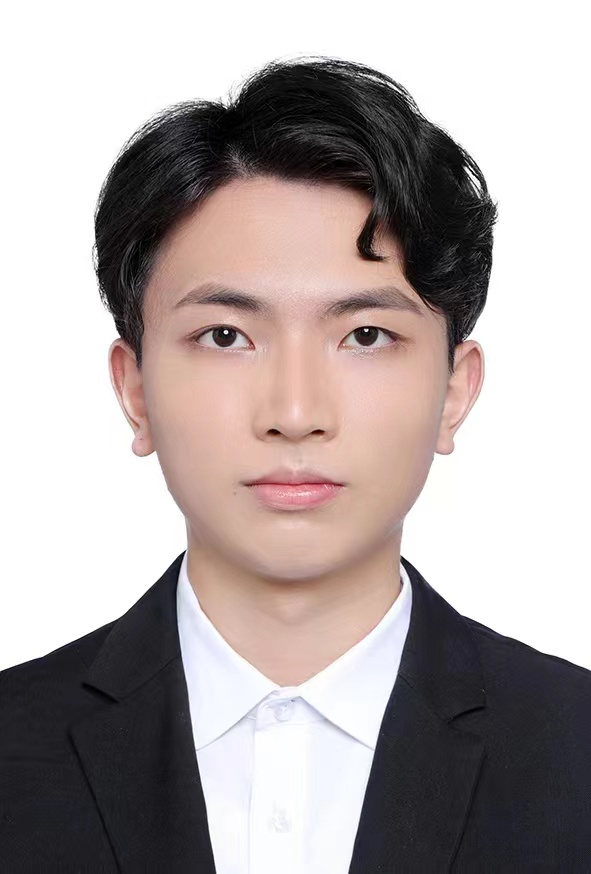}}]{Junfei Wu}
is currently pursuing his Ph.D. degree of Computer Science at the Center for Research on Intelligent Perception and Computing (CRIPAC), State Key Laboratory of Multimodal Artificial Intelligence Systems (MAIS), Institute of Automation, Chinese Academy of Sciences (CASIA). His current research interests mainly include fake news detection.
\end{IEEEbiography}

\begin{IEEEbiography}[{\includegraphics[width=1in,height=1.25in,keepaspectratio]{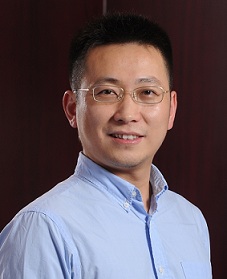}}]{Shu Wu}
received his B.S. degree from Hunan University, China, in 2004, M.S. degree from Xiamen University, China, in 2007, and Ph.D. degree from Department of Computer Science, University of Sherbrooke, Quebec, Canada, all in computer science. He is an Associate Professor with the Center for Research on Intelligent Perception and Computing (CRIPAC), State Key Laboratory of Multimodal Artificial Intelligence Systems (MAIS), Institute of Automation, Chinese Academy of Sciences (CASIA). He has published more than 50 papers in the areas of data mining and information retrieval in international journals and conferences, such as IEEE TKDE, IEEE THMS, AAAI, ICDM, SIGIR, and CIKM. His research interests include data mining, information retrieval, and recommendation.
\end{IEEEbiography}

\begin{IEEEbiography}[{\includegraphics[width=1in,height=1.25in,keepaspectratio]{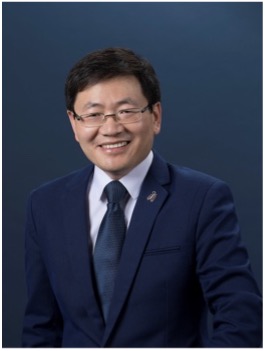}}]{Liang Wang}
received both the BEng and MEng degrees from Anhui University in 1997 and 2000, respectively, and the PhD degree from the Institute of Automation, Chinese Academy of Sciences (CASIA) in 2004. From 2004 to 2010, he was a research assistant at Imperial College London, United Kingdom, and Monash University, Australia, a research fellow at the University of Melbourne, Australia, and a lecturer at the University of Bath, United Kingdom, respectively. Currently, he is a full professor of the Hundred Talents Program at the State Key Laboratory of Multimodal Artificial Intelligence Systems, CASIA. His major research interests include machine learning, pattern recognition, and computer vision. He has widely published in highly ranked international journals such as IEEE TPAMI and IEEE TIP, and leading international conferences such as CVPR, ICCV, and ECCV. He has served as an Associate Editor of IEEE TPAMI, IEEE TIP, and PR. He is an IEEE Fellow and an IAPR Fellow.
\end{IEEEbiography}




\end{document}